\documentclass[10pt,twocolumn,letterpaper]{article}

\usepackage[pagenumbers]{cvpr} 
\usepackage[dvipsnames]{xcolor}
\usepackage{adjustbox}
\usepackage{multirow}
\usepackage{paralist}

\usepackage{graphicx}
\usepackage{amsmath}
\usepackage{amssymb}
\usepackage{booktabs}
\usepackage{pifont}
\newcommand{\cmark}{\ding{51}}%
\newcommand{\xmark}{\ding{55}}%
\usepackage[pagebackref,breaklinks,colorlinks]{hyperref}

\usepackage[capitalize]{cleveref}
\crefname{section}{Sec.}{Secs.}
\Crefname{section}{Section}{Sections}
\Crefname{table}{Table}{Tables}
\crefname{table}{Tab.}{Tabs.}

\def\cvprPaperID{4783} 
\def\confName{CVPR}
\def\confYear{2022}

\newcommand{\gray}[1]{\textcolor{gray}{#1}}
\newcommand{\green}[1]{\textcolor[RGB]{96,177,87}{#1}}
\newcommand{\fn}[1]{\footnotesize{#1}}
\newcommand{\gbf}[1]{\green{\bf{\fn{(#1)}}}}

\newcommand{\pparagraph}[1]{\vspace{1pt} \noindent \textbf{#1} }

\newlength\savewidth

\makeatletter\renewcommand\subsubsection{\@startsection{subsubsection}{4}{\z@}
	{.5em \@plus1ex \@minus.2ex}{-.5em}{\normalfont\normalsize\bfseries}}\makeatother

\usepackage{textpos}  

	\definecolor{Gray}{gray}{0.5}

	\definecolor{Highlight}{HTML}{39b54a}  
	
\begin{document}

\title{ZegFormer: Decoupling Zero-Shot Semantic Segmentation}
\author{
	Jian Ding$^{1,2}$, Nan Xue$^{1}$, Gui-Song Xia$^{1}$\thanks{Corresponding author},
	Dengxin Dai$^{2}$\\
	$^{1}$CAPTAIN, Wuhan University, China \quad $^{2}$MPI for Informatics, Germany\\
	{\tt\small \{jian.ding, xuenan, guisong.xia\}@whu.edu.cn, ddai@mpi-inf.mpg.de}
}
\maketitle

\vspace{-3mm}
\begin{abstract}

Zero-shot semantic segmentation (ZS3) aims to segment the novel categories that have not been seen in the training. 
Existing works formulate ZS3 as a pixel-level zero-shot classification problem, and transfer semantic knowledge from seen classes to unseen ones with the help of language models pre-trained only with texts.
While simple, the pixel-level ZS3 formulation shows the limited capability to integrate vision-language models that are often pre-trained with image-text pairs and currently demonstrate great potential for vision tasks.
Inspired by the observation that humans often perform segment-level semantic labeling, we propose to decouple the ZS3 into two sub-tasks: 1) a class-agnostic grouping task to group the pixels into segments. 2) a zero-shot classification task on segments. The former task does not involve category information and can be directly transferred to group pixels for unseen classes. 
The latter task performs at segment-level and provides a natural way to leverage large-scale vision-language models pre-trained with image-text pairs (\eg CLIP) for ZS3.
Based on the decoupling formulation, we propose a simple and effective zero-shot semantic segmentation model, called ZegFormer, which outperforms the previous methods on ZS3 standard benchmarks by large margins, \eg, \textbf{22 points} on the PASCAL VOC and \textbf{3 points} on the COCO-Stuff in terms of mIoU for unseen classes. Code will be released at \url{https://github.com/dingjiansw101/ZegFormer}.
\end{abstract}

\vspace{-3mm}
\section{Introduction}

\label{sec:intro}
\begin{figure}[!t]
	\centering
	\vspace{-3mm}
	\includegraphics[width=0.99\linewidth]{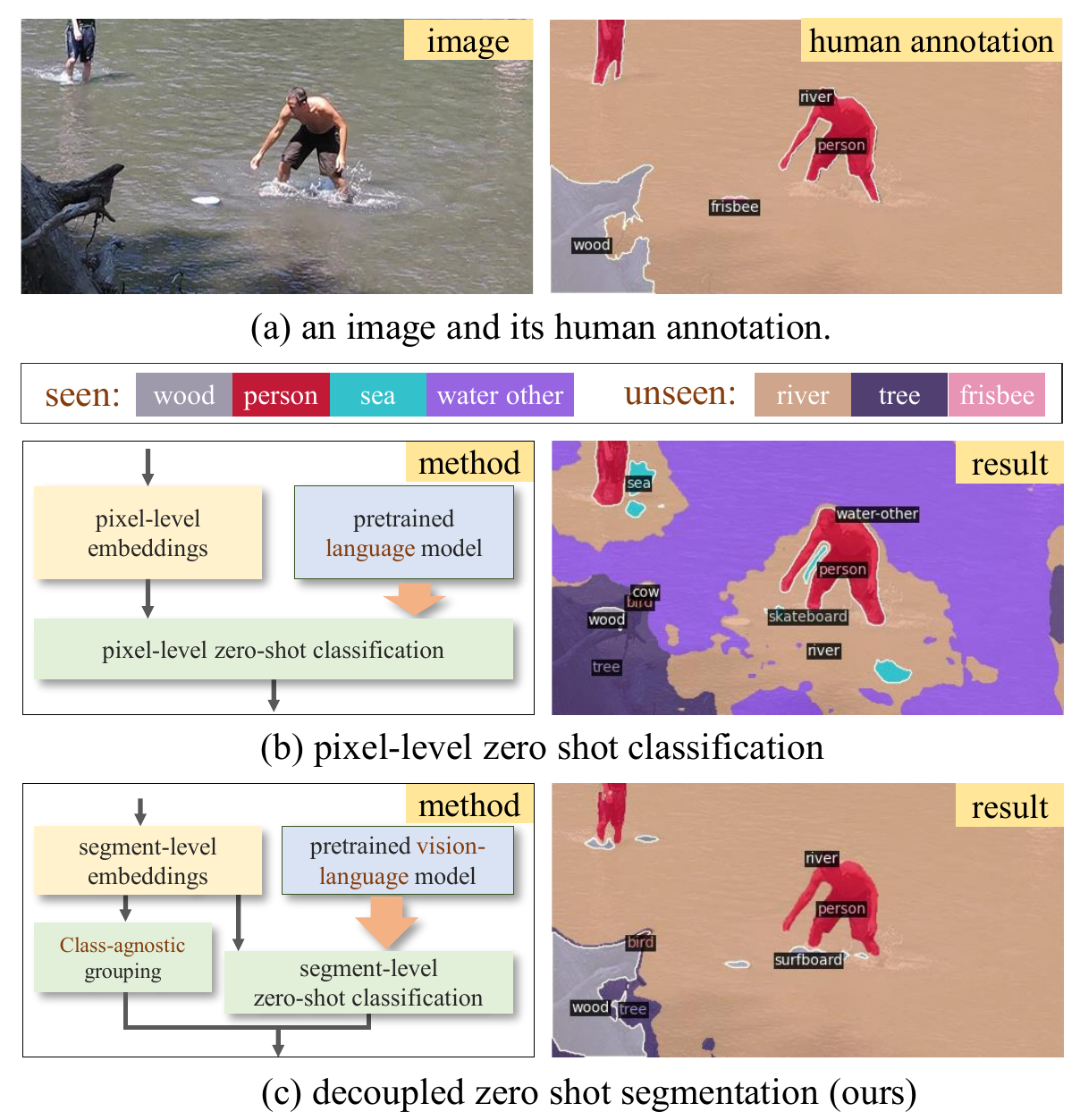}
	\vspace{-2mm}	
	\caption{ 
	ZS3 aims to train a model merely on \textit{seen classes} and generalize it to classes that have not been seen in the training (\textit{unseen classes}). Existing methods formulate it as a pixel-level zero-shot classification problem {\bf (b)}, and use semantic features from a \textit{language} model to transfer the knowledge from seen classes to unseen ones. In contrast, as in {\bf (c)}, we decouple ZS3 into two sub-tasks: 1) A \textit{class-agnostic} grouping and 2) A \textit{segment-level} zero-shot classification, which enables us to take full advantage of the pre-trained \textit{vision-language} model.}
	\label{fig:confusion}
	\vspace{-2mm}
\end{figure}
Semantic segmentation targets to group an image into segments with semantic categories. Although remarkable progress has been made~\cite{FCN,Deeplab,Deeplabv3Plus,pspnet,ocrnet,segformer}, current semantic segmentation models are mostly trained in a supervised manner with a fixed set of predetermined semantic categories, and often require hundreds of samples for each class. 
In contrast, humans can distinguish at least 30,000 basic categories~\cite{biederman1987recognition,fu2017recent}, and  recognize novel categories merely from some \textit{high-level descriptions}. How to achieve human-level ability to recognize stuff and things in images is one of the ultimate goals in computer vision. 

Recent investigations on zero-shot semantic segmentation (ZS3)~\cite{spnet,ZS3Net} have actually moved towards that ultimate goal. Following the fully supervised semantic segmentation models~\cite{FCN,Deeplabv3Plus,Deeplab} and zero-shot classification models~\cite{lampert2013attribute,zhang2016zero,jayaraman2014zero,rohrbach2011evaluating,akata2015evaluation}, these works formulate zero-shot semantic segmentation as a \textit{pixel-level zero-shot classification problem}.
{Although these studies have reported promising results, two main issues still need to be addressed : (1) They usually transfer knowledge from seen to unseen classes by \textit{language} models~\cite{spnet,ZS3Net,fasttext} pre-trained only by texts, which limit their performance on vision tasks. Although large-scale pre-trained \textit{vision-language} models (\eg CLIP~\cite{clip} and ALIGN~\cite{align}) have demonstrated potentials on \textit{image-level} vision tasks, how to efficiently integrate them into the \textit{pixel-level} ZS3 problem is still unknown. 
(2) They usually build correlations between \textit{pixel-level} visual features and semantic features for knowledge transfer, which is \textit{not natural} since we humans often use \textit{words or texts} to describe \textit{objects/segments} instead of \textit{pixels} in images.
As illustrated in Fig.~\ref{fig:confusion}, it is unsurprising to observe that the \textit{pixel-level} classification has poor accuracy on \textit{unseen classes}, which in turn degrades the final segmentation quality. This phenomenon is particularly obvious when the number of unseen categories is large (see Fig.~\ref{fig:visadeinfercoco}).
} 

An intuitive observation is that, given an image for semantic segmentation, we humans can first \textit{group pixels into segments} and then perform a \textit{segment-level semantic labeling} process. For example, a child can easily group the pixels of an object, even though he/she does not know the name of the object. Therefore, we argue that {\em a human-like zero-shot semantic segmentation procedure should be decoupled into two sub-tasks}: 
\begin{itemize}
    \item[-] A {\em class-agnostic grouping} to group pixels into segments. This task is actually a classical image partition/grouping problem~\cite{DDMCMC,NormalizeCUT,supervisedeval}, and can be renewed via deep learning based methods~\cite{maskformer,KNet,panopticsegformer}.
    \item[-] A {\em segment-level zero-shot classification} to assign semantic labels either seen or unseen to segments.
\end{itemize}
As the the grouping task does not involve the semantic categories, a grouping model learned from seen classes can be easily transferred to unseen classes. The segment-level zero-shot classification is robust on the unseen classes and provides a flexible way to integrate the pre-trained large-scale vision-language models~\cite{clip} to the ZS3 problem.
To instantiate the decoupling idea, we present a simple yet efficient {\em zero-shot semantic segmentation model with transformer}, named ZegFormer, which uses a transformer decoder to output \textit{segment-level} embeddings, as shown in Fig.~\ref{fig:pipeline}. It is then followed by a \textit{mask projection} for \textit{class-agnostic grouping} (CAG) and a \textit{semantic projection} for \textit{segment-level zero-shot classification} (s-ZSC). The \textit{mask projection} maps each \textit{segment-level} embedding to a mask embedding, which can be used to obtain a binary mask prediction via a dot product with a high-resolution feature map. The \textit{semantic projection} establishes the correspondences between \textit{segment-level} embedding and semantic features of a pre-trained \textit{text encoder} for s-ZSC. 

While the steps mentioned above can form a standalone approach for ZS3, the model trained on a small dataset is struggling to have strong generalization. Thanks to the decoupling formulation, it is also flexible to use an \textit{image encoder} of a vision-language model to generate image embeddings for zero-shot segment classification. As we empirically find that the segment classification scores with image embeddings and s-ZSC are complementary. We fuse them to achieve the final classification scores for segments.
The proposed ZegFormer model has been extensively evaluated with experiments and demonstrated superiority on various commonly-used benchmarks for ZS3. It outperforms the state-of-the-art methods by \textbf{22 points} in terms of mIoU for unseen classes on the PASCAL VOC~\cite{voc}, and \textbf{3 points} on the  COCO-Stuff~\cite{coco-stuff}. 
Based on the challenging ADE20k-Full dataset~\cite{ade20k}, we also create a new ZS3 benchmark with {\bf 275 unseen classes}, the number of unseen classes in which are much larger than those in PASCAL-VOC ($5$ unseen classes) and COCO-Stuff ($15$ unseen classes). On the ADE20k-Full ZS3 benchmark, our performance is comparable to MaskFormer~\cite{maskformer}, a fully supervised semantic segmentation model. 

Our contributions in this paper are three-fold:
\begin{itemize}
	\item 
    We propose a new formulation for the task of ZS3, by decoupling it into two sub-tasks, {\em a class-agnostic grouping} and {\em a segment-level zero-shot classification}, which provides a more natural and flexible way to integrate the pre-trained large-scale vision-language models into ZS3. 

	\item  With the new formulation, we present a simple and effective ZegFormer model for ZS3, which uses a transformer decoder to generate \textit{segment-level} embeddings for grouping and zero-shot classification. To the best of our knowledge, the proposed ZegFormer is the first model taking full advantage of the pre-trained large-scale vision-language model (\eg CLIP~\cite{clip}) for ZS3.
	\item We achieved state-of-the-art results on standard benchmarks for ZS3. The ablation and visualization analyses show that the decoupling formulation is superior to pixel-level zero-shot classification by a large margin.

\end{itemize}

\section{Related Works}

\paragraph{Zero-Shot Image Classification}~aims to classify images of \textit{unseen categories} that have not been seen during training. The key idea in zero-shot learning is to transfer kowledge from \textit{seen classes} to \textit{unseen classes} via semantic representation, such as the semantic attributes~\cite{lampert2013attribute,al2016recovering,jayaraman2014zero,kankuekul2012online}, concept ontology~\cite{miller1995wordnet,fergus2010semantic,rohrbach2011evaluating,rohrbach2010helps,mensink2014costa} and semantic word vectors~\cite{Devise,norouzi2013zero,zhang2016zero}.
Recently, there are some works that use large-scale \textit{vision-language pretraining}~\cite{clip,align} via contrastive loss. For example, by training a vision-language model on 400 million image and text pairs collected by Google Engine \textit{without any human annotations}, CLIP~\cite{clip} has achieved impressive performances on more than 30 vision datasets even compared to the supervised models. It has also shown the potential for zero-shot object detection~\cite{ViLD}. However, there is a large gap between the \textit{pixel level features} used in previous ZS3 models and \textit{image-level features} used in CLIP. To bridge this gap, we build the correspondence between \textit{segment-level} visual features and \textit{image-level} vision-language features.


\vspace{-3mm}
\paragraph{Zero-Shot Segmentation}~is a relatively new research topic~\cite{zhao2017open,ji2018end,spnet,ZS3Net,STRICT}, which aims to segment the classes that have not been seen during training. 
There have been two streams of work: discriminative methods~\cite{spnet,STRICT,baek2021exploiting} and generative methods~\cite{ZS3Net,CaGNet,shen2021conterfactual}.
SPNet~\cite{spnet} and ZS3Net~\cite{ZS3Net} are considered the representative examples. In detail, SPNet~\cite{spnet} maps each pixel to a semantic word embedding space and projects each pixel feature into class probability via a fixed semantic word embedding~\cite{fasttext,word2vec} projection matrix. ZS3Net~\cite{ZS3Net} first train a generative model to generate pixel-wise features of unseen classes by word embeddings. With the synthetic features, the model can be trained in a supervised manner. Both of these two works formulate ZS3 as a \textit{pixel-level zero-shot classification} problem. However, this formulation is not robust for ZS3, since the text embeddings are usually used to describe objects/segments instead of pixels. The later works~\cite{CSRL,CaGNet,CaGNetv2,kendall2017uncertainties,shen2021conterfactual,cap2seg} all follow this formulation to address different issues in ZS3. 
In a weaker assumption that the unlabelled pixels from unseen classes are available in the training images, self-training~\cite{ZS3Net,STRICT} is widely used. Although promising performances are reported, self-training often needs to retrain a model whenever a new class appears.
Different from the pixel-level zero-shot classification, we propose a new formulation for ZS3, by decoupling ZS3 into a class-agnostic learning problem on pixels and a segment-level zero-shot learning problem. Then we implement a ZegFormer for ZS3, which does not have a complicated training scheme and is flexible to transfer to new classes without retraining. A recent work~\cite{zeroshotinst} also uses region-level classification for \textit{bounding boxes}. But it focuses on \textit{instance segmentation} instead of \textit{semantic segmentation}.  Besides, it still predicts class-\textit{\textbf{aware}} masks. We are the first to use the region-level classification for \textit{zero-shot semantic segmentation}.

\vspace{-6mm}
\paragraph{Class-Agnostic Segmentation} is a long-standing problem and has been extensively studied in computer vision~\cite{supervisedeval,DDMCMC,NormalizeCUT,MCG,UCM,deng1999color,zhu1996region}. There are evidences~\cite{maskrcnn,everything,deepmask} that \textit{class-agnostic} segmentation model learned from seen classes can be well transferred to \textit{unseen classes} in the task of \textit{instance segmentation}, under a partially supervised training paradigm. Recently, a class-agnostic segmentation task~\cite{entityseg} called \textit{entity segmentation (ES)} is proposed, which can predict segments for both thing and stuff classes. However, ES is an \textit{instance-aware task}, which is different from \textit{semantic segmentation}. In addition, entity segmentation~\cite{entityseg} does not predict the detailed class names of unseen classes. Our work is inspired by the abovementioned findings, but we focus on the \emph{semantic segmentation} of novel classes and also predict the detailed class names of unseen classes. With the formulation, our proposed method is simpler, flexible, and robust.

\section{Methodology}
\subsection{Decoupling Formulation of ZS3 }\label{sec:formulation}
Given an image $I$ on the domain $\Omega = \{0,\ldots,H-1\}\times \{0,\ldots,W-1\}$, 
the semantic segmentation of $I$ can be defined as a process to find a pair of mappings $(\mathcal{R}, \mathcal{L})$ for $I$, where $\mathcal{R}$ groups the image domain into $N$ piece-wise ``homogeneous'' segments $\{R_i\}_{i=1}^N$, such that $\cup^N_{i=1} R_i = \Omega$ and $R_i \cap R_j = \O$, if $i \neq j$, and $\mathcal{L}$ associates every segment $R_i \in \Omega$ with a semantic label $c\in C$, where $C$ is a predefined set of categories. 

A fully-supervised semantic segmentation suggests that, one can learn such a pair of mappings $(\mathcal{R}, \mathcal{L})$ for $I$ from a large-scale semantic annotated dataset, \ie, $\mathbf{D} = \{I_k, \mathcal{R}_k, \mathcal{L}_k\}_{k=1}^K$. This type of methods are often with an assumption that the category set $C$ are closed, \ie, the categories appearing in testing images are well contained by $C$, which, however are usually violated in real-application scenarios. 

Actually, if we denote $S$ as the category set of the annotated dataset $\mathbf{D}$, \ie, the seen classes, and $E$ as those appearing in the testing process, we have three types of settings for semantic segmentation, 
\begin{itemize}
\item fully-supervised semantic segmentation: $E\subseteq S$,
\item zero-shot semantic segmentation (ZS3): $S\cap E = \O$,
\item generalized ZS3 (GZS3) problem: $S \subset E$.
\end{itemize} 
In this paper, we mainly address the problem of GZS3, and denote $U = E - E\cap S$ as the set of unseen classes.

\pparagraph{Relations to Pixel-Level Zero-Shot Classification.} 
Previous works~\cite{spnet,ZS3Net,CaGNet} formulate ZS3 as a \textit{pixel-level zero-shot classification} problem, where a model learned from pixel-level semantic labels of $S$ needs to be generalized to pixels of $U$.
It can be considered as \textit{a special case} of our decoupled formulation, where each pixel represents a segment $R_i$. 
Since the learning of $\mathcal{R}$ does not involve the semantic categories, our formulation separates a \textit{class-agnostic learning sub-task} from ZS3. The \textit{class-agnostic} task has a strong generalization to unseen categories, as demonstrated in~\cite{deepmask,entityseg}. 
In addition, as humans often associate semantics to whole images or at least segments, establishing a connection from semantics to \textit{segment-level} visual features is more \textit{natural} than that to \textit{pixel-level} visual features. Therefore, our formulation is more efficient than \textit{pixel-level zero-shot classification} in transferring knowledge from $S$ to $U$.



\subsection{ZegFormer}
\begin{figure*}[!t]
	\centering
	\vspace{-3mm}
	\includegraphics[width=0.9\linewidth]{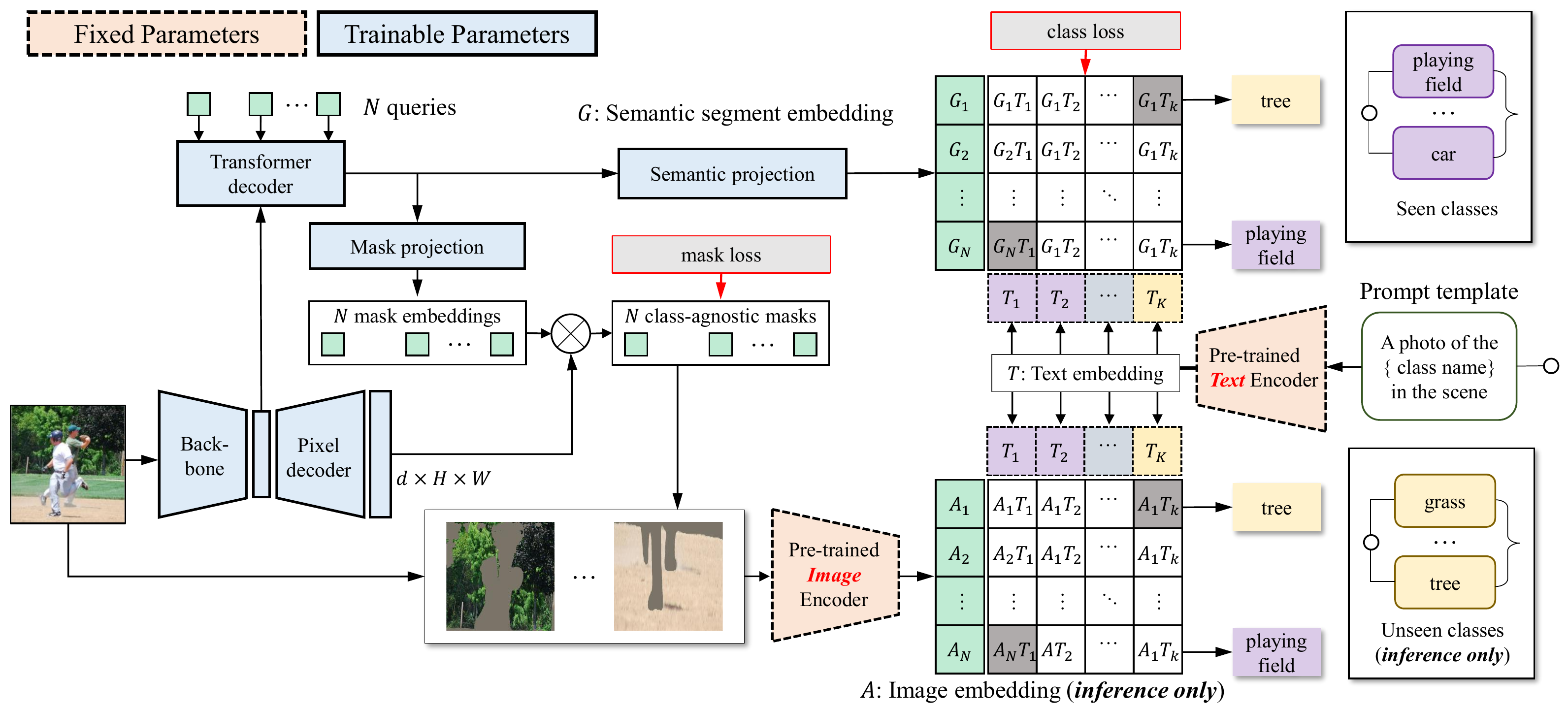}
	\vspace{-3mm}	
	\caption{The pipeline of our proposed ZegFormer for zero-shot semantic segmentation. We first feed $N$ queries and feature maps to a transformer decoder to generate $N$ \textit{segment embeddings}. We then feed each segment embedding to a mask projection layer and a semantic projection layer to obtain a mask embedding and a semantic embedding. Mask embedding is multiplied with the output of pixel decoder to generate a \textit{class-agnostic binary mask}, while The semantic embedding is classified by the text embeddings. The text embeddings are generated by putting the class names into a prompt template and then feeding them to a text encoder of a vision-language model. During training, only the seen classes are used to train the segment-level classification head. During inference, both the text embeddings of \textit{seen} and \textit{unseen} classes are used for segment-level classification. We can obtain two segment-level classification scores with semantic segment embeddings and image embeddings. Finally, we fuse the these two classification scores as our final class prediction of segments. 
	}
	\label{fig:pipeline}
	\vspace{-3mm}
\end{figure*}

Fig.~\ref{fig:pipeline} illustrates the pipeline of our proposed ZegFormer model. 
We first generate a set of segment-level embeddings and then project them for class-agnostic grouping and segment-level zero-shot classification by two parallel layers. A pre-trained image encoder is also used for segment classification. The two segment-level classification scores are finally fused to obtain the results.
\paragraph{Segment Embeddings.}
Recently, there are several segmentation models~\cite{maskformer,panopticsegformer,KNet} that can generate a set of \textit{segment-level} embeddings. We choose the Maskformer~\cite{maskformer} as a basic semantic segmentation model for simplicity. By feeding $N$ segment queries and a feature map to a transformer decoder, we can obtain $N$ segment-level embeddings. 
Then we pass each segment embedding through a semantic projection layer and a mask projection layer to obtain a \textit{mask embedding} and a \textit{semantic embedding} for each segment. We denote segment-level \textit{semantic embedding} (SSE) as $G_q\in \mathbb{R}^{d}$ and the segment-level \textit{mask embedding} as $B_q\in \mathbb{R}^{d}$, where $q$ indexes a query.  
\vspace{-4mm}
\paragraph{Class-Agnostic Grouping.}
Denote the feature maps out from pixel decoder as $\mathcal{F}(I)\in \mathbb{R}^{d\times H \times W}$. The binary mask prediction for each query can be calculated as $m_q = \sigma(B_q \cdot \mathcal{F}(I))\in [0,1]^{H\times W}$, where $\sigma$ is the sigmoid function. Note that $N$ is usually smaller than the number of classes.
\vspace{-4mm}
\paragraph{Segment Classification with SSE.}
For training and inference, each ``class name" in a set of class names $C$ is put into a prompt template (\eg ``A photo of a \{class name\} in the scene") and then fed to a text encoder. 
Then we can obtain $|C|$ text embeddings, denoted as $T = \{T_c\in \mathbb{R}^d |c=1,...|C|\} $, where $C=S$ during training, while $C = S\cup U$ during inference. In our pipeline, we also need a ``no object" category, if the intersection over union (IoU) between a segment and any ground truths is low. For the ``no object" category, it is unreasonable to be presented by a single class name. Therefore, we add an extra learnable embedding $T_0\in \mathbb{R}^{d}$ for ``no object". 
The predicted probability distribution over the seen classes and the ``no object" for a segment query is calculated as:

\begin{equation}
\label{eq:prob} 
p_q(c) = \frac{\exp(\frac{1}{\tau}s_{c} (T_{i}, G_q))}{\sum_{i=0}^{|C|}\exp(\frac{1}{\tau}s_{c}(T_i, G_q))},     
\end{equation}
where $q$ indexes a query. $s_{c}(\bf e, \bf e^{'})=\frac{\bf e\cdot \bf e^{'}}{|\bf e| |\bf e^{'}|}$ is the cosine similarity between two embeddings. $\tau$ is the temperature. 
\vspace{-4mm}
\paragraph{Segment Classification with Image Embedding.}
\label{sec:zerocls}
While the aforementioned steps can already form \textit{a standalone approach} for ZS3, it is also possible to use an 
\textit{image encoder} of a pre-trained vision-language model (\eg CLIP~\cite{clip}) to improve the classification accuracy on segments, owing to the flexibility of the decoupling formulation.
In this module, we create a suitable sub-image for a segment. The process can be formulated as, given a mask prediction $m_q\in [0, 1]^{H\times W}$ for a query $q$, and the input image $I$, create a sub-image $I_q=f(m_q, I)$, where $f$ is a preprocess function (\eg, a masked image or a cropped image according to $m_q$). We give detailed ablation studies in Sec.~\ref{preprocess}. We pass $I_q$ to a pre-trained image encoder and obtain image embedding $A_q$. Similar to Eq.~\ref{eq:prob}, we can calculate a probability distribution, denoted as $p^{'}_{q}(c)$. 



\pparagraph{Training.} 
During the training of ZegFormer, only the pixel labels belonging to $S$ are used.
To compute the training loss, a bipartite matching~\cite{detr,maskformer} is performed between the predicted masks and the ground-truth masks. The loss of classification for each segment query is $-log(p_q(c_q^{gt}))$, where $c_q^{gt}$ belongs to $S$ if the segment is matched with a ground truth mask and ``no object" if the segment does not have a matched ground truth segment. For a segment matched with ground truth segment $R_q^{gt}$, there is a mask loss $L_{mask}(m_q, R_q^{gt}) $. In detail, we use the combination of a dice loss~\cite{dice} and a focal loss~\cite{focalloss}.

\pparagraph{Inference.}
 During inference, we integrate the predicted binary masks and class scores of segments to obtain the final results of semantic segmentation. According to the class probability scores, we have three variants of ZegFormer.
 
\vspace{.5em}\noindent
(1) \emph{ZegFormer-seg.} This variant use the segment classification scores with segment queries for inference by calculate a per-pixel class probability $\sum_{q=1}^{N}p_q(c)\cdot m_q[h,w]$, where $(h,w)$ is a location in an image. Since there is an imbalanced data problem, which results in predictions being biased to seen classes. Following~\cite{spnet}, we calibrate the prediction by decreasing the scores of seen classes. The final category prediction for each pixel is then calculated as:
\begin{equation}
\label{eq:inference}
    \mathop{\arg\max}_{c\in S+U} \sum_{q=1}^{N}p_q(c)\cdot m_q[h,w] - \gamma \cdot \mathbb I[c\in S], 
\end{equation}
where $\gamma \in [0,1]$ is the calibration factor. The indicator function $\mathbb I$ is equal to $1$ when $c$ belongs to the seen classes.

\vspace{.5em}\noindent
(2) \emph{ZegFormer-img.} The inference process of this variant is similar to Eq.~\ref{eq:inference}. The only difference is that the $p_q(c)$ is replaced by $p_q^{'}(c)$.

\vspace{.5em}\noindent
(3) \emph{ZegFormer.} This variant is our full model. We first fuse $p_q(c)$ and $p^{'}_q(c)$ for each query as:
\begin{equation}
    p_{q, \text{fusion}}(c) = \begin{cases}
    p_{q}(c)^\lambda \cdot  p_{q, \text{avg}}^{(1-\lambda)} & \text{if } c \in S \\
    p_{q}(c)^{(1-\lambda)} \cdot p_{q}^{'}(c)^\lambda & \text{if } c \in U, 
\end{cases} 
\label{eq:ensemble}
\end{equation}
where a geometry mean of $p_q(c)$ and $p_q^{'}(c)$ will return, if $c\in U$. The contribution of the two classification scores is balanced by $\lambda$.  
Since $p_q(c)$ is usually more accurate than $p^{'}_q(c)$ if $c\in S$, we do not want $p_q^{'}(c)$ contribute to the prediction of $S$. 
Therefore, we calculate a geometry mean of $p_q(c)$ and $p_{q, \text{avg}} = \sum_{j\in S} p_{q}^{'}(j) / |S|$ on seen classes.
This way, the probabilities of seen classes and unseen classes can be adjusted to the same range, and only $p_q(c)$ contributes to distinguishing seen classes. The final results for semantic segmentation are obtained by a process similar to Eq.~\ref{eq:inference}.
\section{Experiments}
Since most of the previous works~\cite{CaGNet,ZS3Net,STRICT} focus on the GZS3 setting, we evaluate our method on the GZS3 setting in the main paper. See our results on the ZS3 setting in the appendix. We introduce the datasets and evaluation metrics that we use in the following.
\subsection{Datasets and Evaluation Metrics}
\pparagraph{COCO-Stuff} is a large-scale dataset for semantic segmentation that contains 171 valid classes in total. We use 118,287 training images as our training set and 5,000 validation images as our testing set. We follow the class split in~\cite{spnet} to choose 156 valid classes as seen classes and 15 valid classes as unseen classes. We also use a subset of the training classes as a validation set for tuning hyperparameters, following the cross-validation procedure~\cite{spnet,xian2018zero}.

\pparagraph{PASCAL-VOC Dataset} has been split into 15 seen classes and 5 unseen classes in previous works~\cite{spnet,CaGNet}.
There are 10582 images for training and 1,449 images for testing. 

\pparagraph{ADE20k-Full Dataset} is annotated in an open-vocabulary setting with more than 3,000 categories. It contains 25k images for training and 2k images for validation. We are the first that evaluate GZS3 methods on the challenging ADE20k-Full. Following the supervised setting in ~\cite{maskformer}, we choose 847 classes that are present in both train and validation sets for evaluation, so that we can compare our ZegFormer with supervised models. We split the classes into seen and unseen according to their \textit{frequency}. The seen classes are present in more than 10 images, while the unseen classes are present in less than 10 images. In this way, we obtain 572 classes for seen and 275 classes for unseen.


\pparagraph{Class-Related Segmentation Metric} includes the \textit{mean intersection-over-union} (mIoU) averaged on seen classes, unseen classes, and their harmonic mean, which has been widely used in previous works of GZS3~\cite{spnet,ZS3Net}.

\pparagraph{Class-Agnostic Grouping Metric} has been well studied in ~\cite{supervisedeval}. We use the well-known \textit{precision-recall for boundaries} ($P_b$, $R_b$, and $F_b$) as the evaluation metric, and use the public available code\footnote{\url{https://github.com/jponttuset/seism}} for evaluation.

\subsection{Implementation Details}
Our implementation is based on Detectron2~\cite{wu2019detectron2}.
For most of our ablation experiments, we use ResNet50 as our backbone and FPN~\cite{FPN} as the pixel decoder. When compared with the state-of-the-art methods, we use the ResNet-101 as our backbone. We use the text encoder and image encoder of ViT-B/16 CLIP model\footnote{\url{https://github.com/openai/CLIP}} in our implementation. We set the number of queries in the transformer decoder as 100 by default. The dimension of query embedding and transformer decoder is set as 256 by default. Since the dimension of text embeddings is 512, we use a projection layer to map the segment embeddings from 256 to 512 dimension. We empirically set the temperature $\tau$ in Eq.~\ref{eq:prob} to be 0.01. The image resolution of processed sub-images for segment classification with image-level embeddings is $224\times 224$. 
For all the experiments, we use a batch size of 32.  We train models for 60k and 10k iterations for COCO-Stuff and PASCAL VOC, respectively. We use the ADAMW as our optimizer with a learning rate of 0.0001 and 1e-4 weight decay. 

\vspace{-1mm}
\subsection{Zero-Shot Pixel Classification Baseline}
To compare \textit{decoupling formulation} with \textit{pixel-level zero-shot classification}, we choose SPNet~\cite{spnet} as our \textit{pixel-level zero-shot classification} baseline, since SPNet and ZegFormer both belong to the \textit{discriminative} methods with neat designs. For fair comparison, we reimplement the SPNet~\cite{spnet} with FPN (which is also used by ZegFormer). We denote this variant of SPNet as SPNet-FPN. 
We implement SPNet-FPN in the same codebase (\textit{i.e.}, Detectron2~\cite{wu2019detectron2}) as our ZegFormer. The common settings are also the same as the ones used in ZegFormer). 

\subsection{Comparisons with the Baseline}
\pparagraph{Class-Related Segmentation Metric.}
We compare ZegFormer-seg with the baseline under two types of text embeddings (\textit{i.e.} CLIP~\cite{clip} text embeddings and the concatenation of fastText~\cite{fasttext} (ft) and word2vec~\cite{word2vec} (w2v).) 
The widely used ft + w2v in GZS3\cite{spnet} is trained only by \textit{language data}. In contrast, CLIP text encoder is trained by multimodal contrastive learning of \textit{vision and language}. From Tab.~\ref{tab:class_embeddings}, we can conclude: 
1) CLIP text embeddings~\footnote{Similar to works in object detection~\cite{openvocdet,ViLD}, when using CLIP text embeddings, it is a relaxed setting compared to the pure GZS3 task. But this setting has a more realistic value. For simplicity, we still call it GZS3.} is better than the concatenation of fastText~\cite{fasttext} and word2vec~\cite{word2vec}; 2) Our proposed decoupling formulation for ZS3 is better than the commonly-used zero-short pixel classification ones, regardless of what class embedding methods are used; 3) \textit{ZegFormer-seg has a much large gain than the SPNet-FPN}, when the class embedding method is changed from the ft + w2v to CLIP (\textbf{10.6} points \textit{v.s.} \textbf{4.1} points improvements). We argue that the segment-level visual features are aligned better to the features of CLIP than the pixel-level visual features.
\begin{table}[!t]
\centering
\small
\caption{Comparisons with the baseline in class-related segmentation metric on COCO-stuff. We report the results with CLIP~\cite{clip} text embeddings and the concatenation of fastText (ft)~\cite{fasttext} and word2vec (w2v)~\cite{word2vec} for each algorithm. We show the improvements from ft+w2v to CLIP text in brackets.}
\vspace{-3mm}
\resizebox{\linewidth}{!}{
\begin{tabular}{c|c|lll}
\toprule

                               &             & \multicolumn{3}{c}{Generalized zero shot} \\ \hline
                               & cls. embed. & Seen       & Unseen       & Harmonic      \\ \hline
\multirow{2}{*}{SPNet-FPN}     & ft+w2v      & 31.8       & 6.9          & 11.3          \\
                               & clip text   & 32.3~\gbf{+0.5}       & 11.0~\gbf{+4.1}         & 16.4~\gbf{+5.1}          \\ \hline
\multirow{2}{*}{ZegFormer-seg (ours)} & ft+w2v      & 37.3       & 10.8         & 16.8          \\
                               & clip text   & \textbf{37.4}~\gbf{+0.1}        & \textbf{21.4}~\gbf{+10.6}         & \textbf{27.2}~\gbf{+10.4}          \\ 
\bottomrule
\end{tabular}
}
\label{tab:class_embeddings}
\vspace{-2mm}
\end{table}

\begin{table}[!t]
\small
\centering
\vspace{-1mm}
\caption{Comparisons with the baseline in class-agnostic grouping metric on COCO-Stuff. In the brackets
are the gaps to the SPNet-FPN (baseline).}
\vspace{-3mm}
\resizebox{\linewidth}{!}{
\begin{tabular}{c|c|ccc}
\toprule
              & cls. embed. & $F_b$ & $P_b$ & $R_b$ \\ \hline
SPNet-FPN     & ft+w2v      & 40.3  & 32.6  & 52.8  \\ 
ZegFormer-seg (ours) & ft+w2v      & 49.4~\gbf{+9.1}    & 43.5~\gbf{+10.9}  & 57.2~\gbf{+4.4}  \\ \hline
SPNet-FPN     & clip text   & 42.6  & 35.4  & 53.4  \\
ZegFormer-seg (ours) & clip text   & 50.4~\gbf{+7.8}  & 44.0~\gbf{+8.6}  & 58.9~\gbf{+5.5}  \\
\bottomrule
\end{tabular}
}
\label{partitionquality}
\end{table}
\vspace{-6mm}
\paragraph{Class-Agnostic Grouping Metric.} We compare the image grouping quality of ZegFormer-seg and the baseline on COCO-Stuff~\cite{coco-stuff}. The image grouping quality is a \textit{class-agnostic metric}, which can provide us more insight. We can see in Tab.~\ref{partitionquality} that ZegFormer significantly outperforms the baseline on the $F_b$, $P_b$, and $R_b$, regardless of what class embedding methods are used. This verifies that the \textit{decoupled ZS3 formulation} has much stronger generalization than \textit{pixel-level zero-shot classification} to group the pixels of unseen classes.

\vspace{-3mm}
\begin{table}[!t]
\small
\caption{Influence of preprocess of for sub images.}
\vspace{-3mm}
\resizebox{\linewidth}{!}{
\begin{tabular}{c|c|ccc}
\toprule
                           & preprocess    & Seen & Unseen & Harmonic \\ \hline
ZegFormer-seg              & -          & \textbf{37.4} & 21.4   & 27.2     \\ \hline
\multirow{3}{*}{ZegFormer} & crop          & 36.6 & 19.7   & 25.6     \\
                           & mask          & 36.0 & 31.0   & 33.3     \\
                           & crop and mask & 35.9 & \textbf{33.1}  & \textbf{34.4}     \\ 
\bottomrule
\end{tabular}
\label{tab:clipsegment}
}
\vspace{-3mm}
\end{table}

\begin{table*}[!t]
\small
\vspace{-3mm}
\caption{Comparison with the previous GZS3 methods on PASCAL VOC and COCO-Stuff. The ``Seen", ``Unseen", and ``Harmonic" denote mIoU of seen classes, unseen classes, and their harmonic mean. The STRICT~\cite{STRICT} proposed a self-training strategy and applied it to SPNet~\cite{spnet}.  The numbers of STRICT and SPNet (w/ self-training) are from ~\cite{STRICT}. Other numbers are from their original papers. }
\vspace{-3mm}
\centering
\resizebox{0.9\linewidth}{!}{
\begin{tabular}{c|c|c|c|ccc|ccc}
\toprule
                                &                &            &                               & \multicolumn{3}{c|}{PASCAL VOC} & \multicolumn{3}{c}{COCO-Stuff} \\ \hline
Methods                         & Type           & w/ Self-train. & Re-train. for new classes? & Seen    & Unseen   & Harmonic   & Seen   & Unseen   & Harmonic   \\ \hline
SPNet~\cite{spnet}              & discriminative &  \xmark   & \xmark                       & 78.0    & 15.6     & 26.1       & 35.2   & 8.7      & 14.0       \\
ZS3~\cite{ZS3Net}               & generative     &  \xmark   & \cmark                    & 77.3    & 17.7     & 28.7       & 34.7   & 9.5      & 15.0       \\
CaGNet~\cite{CaGNet}            & generative     &  \xmark   & \cmark                    & 78.4    & 26.6     & 39.7       & 33.5   & 12.2     & 18.2       \\
SIGN~\cite{SIGN}                & generative     &  \xmark   & \cmark                    & 75.4    & 28.9     & 41.7       & 32.3   & 15.5     & 20.9       \\
Joint~\cite{baek2021exploiting} & discriminative &  \xmark   & \xmark                       & 77.7    & 32.5     & 45.9       & -   & -     & -       \\ \hline
ZS3 ~\cite{ZS3Net}              & generative     & \cmark & \cmark                    & 78.0    & 21.2     & 33.3       & 34.9   & 10.6     & 16.2       \\
CaGNet~\cite{CaGNet}            & generative     & \cmark & \cmark                    & 78.6    & 30.3     & 43.7       & 35.6   & 13.4     & 19.5       \\
SIGN~\cite{SIGN}                & generative     & \cmark & \cmark                    & 83.5    & 41.3     & 55.3       & 31.9   & 17.5     & 22.6       \\
SPNet~\cite{STRICT}             & discriminative & \cmark & \cmark                    & 77.8    & 25.8     & 38.8       & 34.6   & 26.9     & 30.3       \\
STRICT~\cite{STRICT}            & discriminative & \cmark & \cmark                    & 82.7    & 35.6     & 49.8       & 35.3   & 30.3     & 32.6       \\ \hline
ZegFormer                       & discriminative & \xmark    & \xmark                       & \textbf{86.4}   & \textbf{63.6}     & \textbf{73.3}       & \textbf{36.6}  & \textbf{33.2}    & \textbf{34.8}      \\
\bottomrule
\end{tabular}
\label{sota}
}
\vspace{-2mm}
\end{table*}

\begin{figure}[!t]
	\centering
	\includegraphics[width=0.97\linewidth]{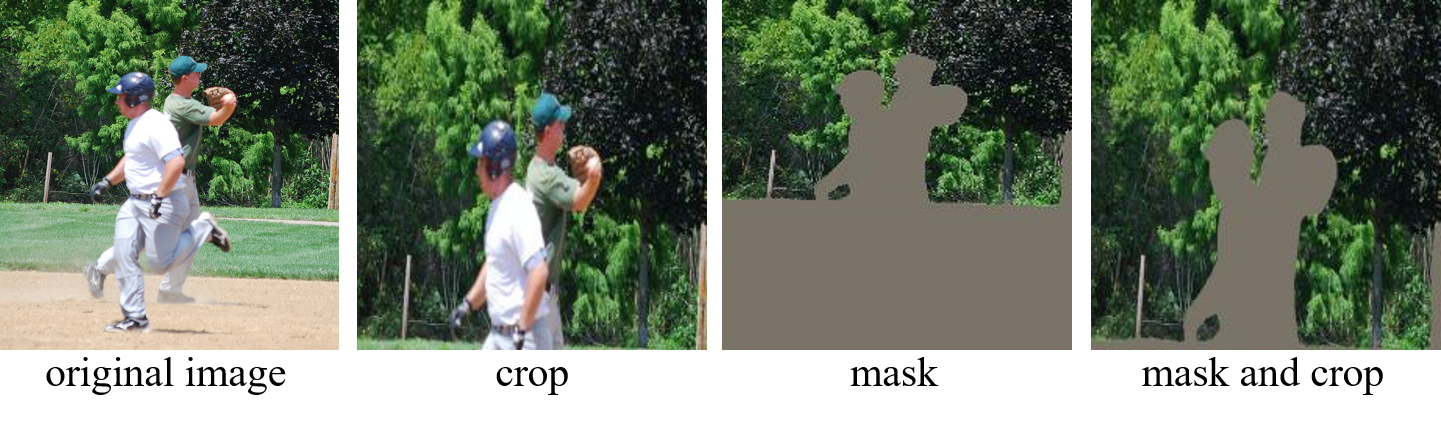}
	\vspace{-4mm}	
	\caption{Comparison between three preprocess for a segment. 
	}
	\label{fig:vis_cropmask}
	\vspace{-2mm}
\end{figure}

\begin{figure}[!t]
	\centering
	\includegraphics[width=0.92\linewidth]{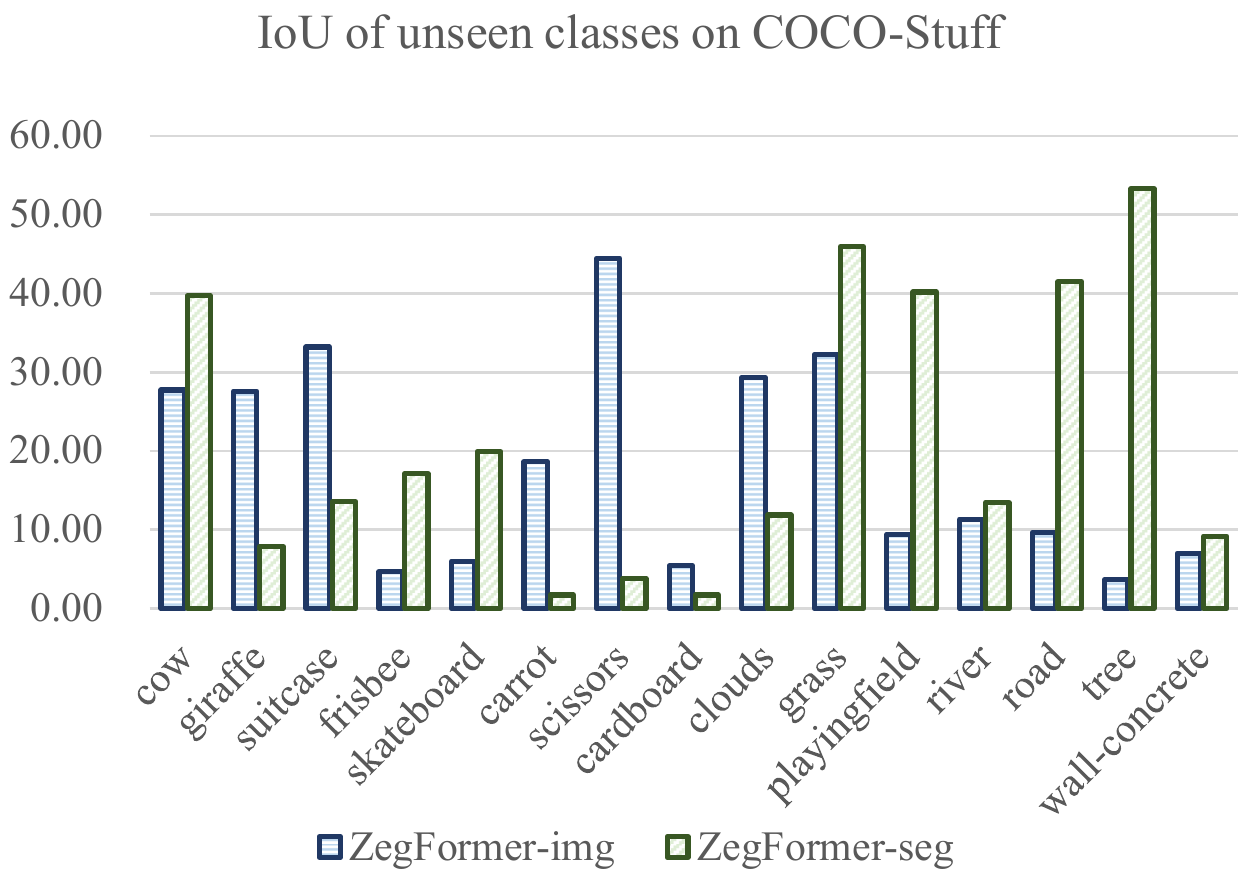}
	\vspace{-3mm}	
	\caption{ZegFormer-seg {\em v.s.} ZegFormer-img in IoU of unseen classes on COCO-Stuff. }
	\label{fig:cls_compare}
\end{figure}

\subsection{Ablation Study on ZegFormer}
\label{sec:ablation}

\paragraph{Preprocess for Sub-Images.}\label{preprocess} 
We explored three preprocess to obtain sub images (\textit{i.e.} ``crop", ``mask", and ``crop and mask") in the full model ZegFormer. 
The three ways to preprocess a segment for ``tree" are shown in Fig.~\ref{fig:vis_cropmask}. We can see that when we merely crop a region from the original image, there may exist more than one categories, which will decrease the classification accuracy. In the masked image, the nuisances are removed, but there are many unnecessary pixels outside the segment. The combination of crop and mask can obtain a relatively suitable image for classification. We compare the influences of three ways for the ZegFormer in Tab.~\ref{tab:clipsegment}. We can see that the combination of crop and mask can get the best performance, while only using the crop preprocess is lower than the performance of ZegFormer-seg, which does not use an image embedding for segment classification. 
\vspace{-3mm}
\paragraph{ZegFormer-seg {\em v.s.} ZegFormer-img.}
We compare their performance for unseen categories on COCO-Stuff in Fig.~\ref{fig:cls_compare}. We can see that the ZegFormer-img is better at things categories (such as ``giraffe", ``suitacase", and ``carrot", etc), while worse at stuff categories (such as ``grass", ``playingfield", and ``river", etc.) 
Therefore, these two kinds of classification scores are complementary, which illustrates why their fusion will improve performance. 
\subsection{Comparison with the State-of-the-art}
We compare our ZegFormer with the previous methods in Tab.~\ref{sota}.
Specifically, ZegFormer outperforms Joint~\cite{baek2021exploiting} by \textbf{31 points} in the mIoU of unseen classes on PASCAL VOC~\cite{voc}, and outperforms SIGN~\cite{SIGN} by \textbf{18 points} in the mIoU of unseen classes on COCO-Stuff~\cite{coco-stuff}. When compared to the results with \textit{self-training}, ZegFormer outperforms SIGN~\cite{SIGN} by \textbf{22 points} in mIoU of unseen classes on PASCAL VOC~\cite{voc}, and STRICT~\cite{STRICT} by \textbf{3 points} on COCO-Stuff~\cite{coco-stuff}.
It is worth noting that the \textit{generative} methods and \textit{self-training} methods need a complicated multi-stage training scheme. They also need to be \textit{re-trained} whenever new classes are incoming (although semantic category labels of unseen classes are not required). The self-training also needs to access the unlabelled pixels of unseen classes to generate pseudo labels. 
 In contrast, our ZegFormer is a \textit{discriminative} method, which is much simpler than those methods and can be applied to any unseen classes \textit{on-the-fly}. Similar to our ZegFormer, SPNet~\cite{spnet} (without self-training) and Joint~\cite{baek2021exploiting} are also discriminative methods that can be flexibly applied to unseen classes, but their performances are much worse than ours.
\begin{figure*}[!t]
	\centering
	\vspace{-3mm}
	\includegraphics[width=0.9\linewidth]{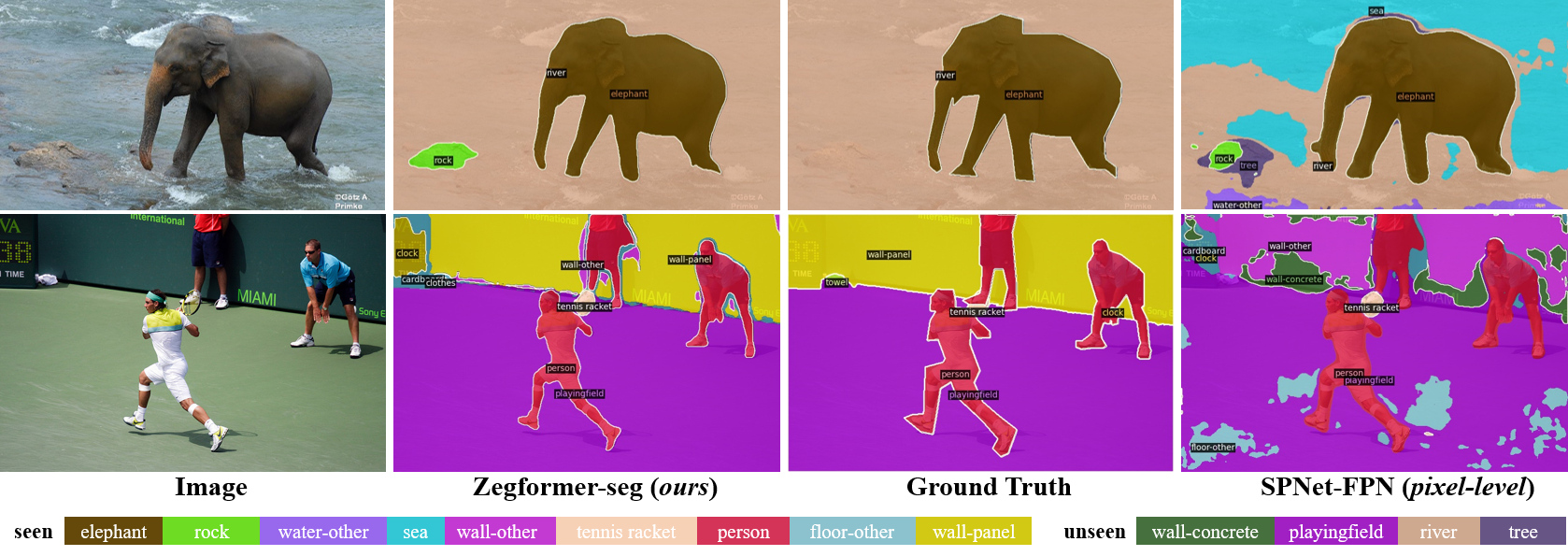}
	\vspace{-3mm}	
	\caption{Results on COCO-Stuff, using 171 class names in COCO-Stuff to generate text embeddings. ZegFormer-seg (our decoupling formulation of ZS3) is better than SPNet-FPN (pixel-level zero-shot classification) in segmenting unseen categories. 
	}
	\label{fig:viscoco}
\end{figure*}

\begin{figure*}[!t]
	\centering
	\vspace{-2mm}
	\includegraphics[width=0.9\linewidth]{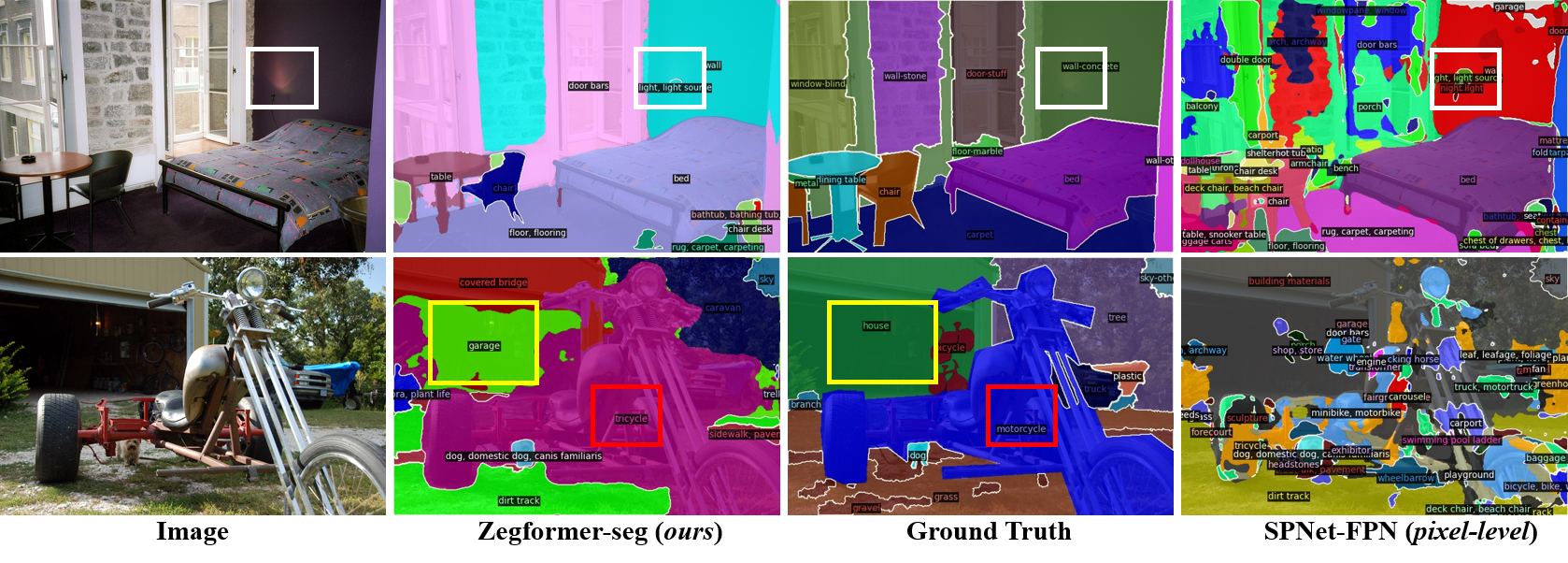}
	\vspace{-3mm}	
	\caption{Results on COCO-Stuff, using 847 class names from ADE20k-Full to generate text embeddings. We can see that the SPNet-FPN (\textit{pixel-level zero-shot classification baseline}) is very unstable when the number of unseen classes is large. We can also see that the set of 847 class names provide richer information than the set of 171 class names in COCO-Stuff. For example, in the yellow box of the second row, the predicted ``garage" for a segment is labeled as ``house" in COCO-Stuff. We provide more visualization results in our appendix.
	}
	\label{fig:visadeinfercoco}
	\vspace{-4mm}
\end{figure*}

\begin{table}[!t]
\small
\caption{Results on ADE20k-Full. Our ZegFormer is comparable with a fully supervised model.}
\vspace{-3mm}
\resizebox{\linewidth}{!}{
\begin{tabular}{c|c|ccc}
\toprule
                           & Backbone & Seen    & Unseen   & Harmonic   \\ \hline
SPNet-FPN             & R-50     & 9.2  & 0.9 & 1.6 \\ 
ZegFormer (ours)                  & R-50     & \textbf{17.4} & \textbf{5.3} & \textbf{8.1} \\ \hline
Fully supervised & R-50     & \gray{19.7} & \gray{5.6} & \gray{8.7} \\ 
\bottomrule
\end{tabular}\label{ade}
}
\vspace{-3mm}
\end{table}

\subsection{Results on ADE20k-Full}
Since we are the first that report the results on the challenging ADE20k-Full~\cite{ade20k} for GZS3, there are no other methods for comparison. We compared ZegFormer with SPNet-FPN, and a fully supervised Maskformer trained on both seen and unseen classes. From Tab.~\ref{ade}, we can see that our ZegFormer significantly surpasses the SPNet-FPN baseline by \textbf{4 points} in mIoU unseen and is \textit{even comparable with the fully supervised model}. We can also see that the dataset is very challenging that even the supervised model only achieved 5.6 points in the mIoU unseen, which indicates there is still much room for improvements. 

\section{Visualization}
For the visualization analyses, we use a SPNet-FPN and a ZegFormer-seg trained with \textbf{156 seen classes} on COCO-Stuff.Then we test the two models with different sets of class names. 
\pparagraph{Results with 171 classes from COCO-Stuff.} From Fig.~\ref{fig:viscoco}, we can see that when segmenting unseen classes, it is usually confused by similar classes. Since the pixels of a segment have large variations, these pixel-level classifications are not consistent. In contrast, decoupling formulation can obtain high-quality segmentation results. Based on the segmentation results, the segment-level zero-shot classification is more related to CLIP pretraining. Therefore, ZegFormer-seg can obtain more accurate classification results.

\pparagraph{Results with 847 classes from ADE20k-Full.} As shown in Fig.~\ref{fig:visadeinfercoco}, the SPNet-FPN (pixel-level zero-shot classification) is much worse than the ZegFormer-seg (our decoupling formulation). The reason is that there is severe competition among classes at pixel-level classification when we use 847 classes for inference. In contrast, the decoupling model is not influenced by the number of classes. These results confirm that \textit{the decoupling formulation is a right way to achieve human-level ability to segment objects with a large number of unseen classes}. We can also see that the set of 847 class names contains richer information to generate text embeddings for inference than the set of 171 class names.
For example, the unannotated ``light, light source" is segmented by the two ZS3 models (white box in the $1_{st}$ row of Fig.~\ref{fig:visadeinfercoco}.) The labeled ``motorcycle" is predicted as ``tricycle" by ZegFormer (red box in the $2_{nd}$ row of Fig.~\ref{fig:visadeinfercoco}). 


\section{Conclusion}
We reformulate the ZS3 task by decoupling. Based on the new formulation, we propose a simple and effective ZegFormer for the task of ZS3, which demonstrates significant advantages to the previous works. The proposed ZegFormer provides a new manner to study the ZS3 problem and serves as a strong baseline. Beyond the ZS3 task, ZegFormer also has the potentials to be used for few-shot semantic segmentation. We leave it for further research.




\pparagraph{Limitations.} {Although the full model of ZegFormer shows superiority in all the situations, we empirically find that ZegFormer-seg does not perform well when the scale of training data is small. One possible reason is that the transformer structure needs a large number of data for training, which has been discussed in ~\cite{liu2021efficient}. A more efficient training strategy for ZegFormer-seg or using other mask classification methods such as K-Net~\cite{KNet} may alleviate this issue, and can be studied in the future.} 

\vspace{-2mm}
\paragraph{Acknowledgement.} 
This work was supported by National Nature Science Foundation of China under grant 61922065, 62101390, 41820104006 and 61871299.
The numerical calculations in this paper have been done on the supercomputing system in the Supercomputing Center of Wuhan University. Jian Ding was also supported by the China Scholarship Council.
We would like to thank Tianzhu Xiang, Di Hu, XingXing Weng and Zeran Ke for insightful discussion. 

\appendix
\section{More Experiments}
\subsection{Results with the ZS3 setting}

Apart from the generalized ZS3 (GZS3), we also report the results achieved with the ZS3 setting, in the evaluation of which the models only predict unseen labels $c\in U$ (see more details in the Sec.~3.1 of our main paper.), and the pixels belong to the seen classes are ignored. The results on COCO-Stuff~\cite{coco-stuff} are reported in Tab.~\ref{tab:coco-stuff}. Since most of the existing studies do not report the results with the ZS3 setting, we compared with the SPNet~\cite{spnet} and re-implemented it with FPN and CLIP~\cite{clip} text embeddings for building a baseline, \ie, SPNet-FPN. The results on ADE20k-Full~\cite{ade20k} are reported by Tab.~\ref{tab:ade20k-full}. 

\begin{table}[h]
	\centering
	\caption{Results on COCO-Stuff with the ZS3 setting. The result of SPNet is directly taken from its original paper~\cite{spnet}, and SPNet-FPN is our re-implementation of the SPNet~\cite{spnet} with FPN and CLIP~\cite{clip} text embeddings, which can be considered as a baseline.}
	\vspace{-3mm}
	\resizebox{\linewidth}{!}{
		\begin{tabular}{c|c|c|c}
			\toprule
			methods          & backbone  & class embed.     & mIoU unseen       \\ \hline
			SPNet~\cite{spnet} & R-101     & ft+w2v           & 35.2             \\ 
			SPNet-FPN          & R-50      &    clip text     & 41.3              \\ \hline
			ZegFormer-seg      & R-50      &    clip text     & 48.8              \\
			ZegFormer          & R-50      &    clip text     & \textbf{61.5}      \\ 
			\bottomrule
		\end{tabular} \label{tab:coco-stuff}
	}
\end{table}

\begin{table}[!t]
	\centering
	\caption{Results on ADE20k-Full achieved with the ZS3 setting. All the models use R-50 as a backbone and CLIP~\cite{clip} as text embeddings.}
	\vspace{-3mm}
	\resizebox{0.57\linewidth}{!}{
		\begin{tabular}{c|c}
			\toprule
			methods              & mIoU unseen                     \\ \hline
			SPNet-FPN              & 7.4                           \\ \hline
			ZegFormer-seg          & 9.4                      \\
			ZegFormer              & \textbf{18.7}                        \\
			\bottomrule
		\end{tabular} 
	}\label{tab:ade20k-full}
\end{table}

\subsection{Speed and Accuracy Analyses}
\pparagraph{Analyses of the Computational Complexity.}
Given $C$ as the number of channels in a feature map, $K$ as the number of classes, $H\times W$ as the size of feature maps that are used for pixel-wise classification, and $N$ being the number of segments in an image, the complexity of the classification head in the \textit{pixel-level zero-shot classification} is $O(H\times W\times C\times K)$, while the complexity of the classification head of our \textit{decoupling formulation} is $O(N\times C\times K)$.
$N$ is usually much smaller than $H\times W$. For instance, in our COCO-Stuff experiments, $N$ is 100, but $H\times W$ is larger than $160\times 160$. Therefore, when $K$ is large, \textit{pixel-level zero-shot classification} will be much slower than the proposed \textit{decoupling formulation of ZS3.} 

\pparagraph{Speed and Accuracy Experiments.}
We compare the speeds of ZegFormer, ZegFormer-\textit{seg}, and the SPNet-FPN. All these three models use R-50 with FPN as a backbone. ZegFormer-\textit{seg} is an implementation for the \textit{decoupling formulation of ZS3}, while SPNet-FPN is our implementation for \textit{pixel-level zero-shot classification}. ZegFormer is our full model, with a branch to generate image embeddings (see Sec.~3.2 of our main paper for details.)
As shown in Tab.~\ref{tab:speed_coco} and Tab.~\ref{tab:speed_ade}, the ZegFormer-\textit{seg} performs better than SPNet-FPN in both speed and accuracy on COCO-Stuff and ADE20k-Full. The ZegFormer improves the ZegFormer-\textit{seg} by \textbf{12 points} in term of mIoU of unseen classes on COCO-Stuff, and still remains an acceptable FPS. We can also see that the speed of SPNet-FPN is slow on ADE20k-Full. This verifies that the speed of \textit{pixel-level zero-shot classification} is largely influenced by $K$ (\textit{number of classes}), as we have discussed before. 
\begin{table}[!t]
	\caption{Results on COCO-Stuff (171 classes) achieved with the GZS3 setting. The FPS is tested on images with the short side of 640 with a single GeForce RTX 3090.}
	\vspace{-3mm}
	\centering
	\resizebox{0.9\linewidth}{!}{
		\begin{tabular}{c|ccc|c}
			\toprule
			methods       & seen & unseen & harmonic & FPS  \\ \hline
			SPNet-FPN     & 32.3 & 11     & 16.4     & 17.0 \\ \hline
			ZegFormer-seg & \textbf{37.4} & 21.4   & 27.2     & \textbf{25.5} \\
			ZegFormer     & 35.9 & \textbf{33.1}   & \textbf{34.4}     & 6.0 \\
			\bottomrule
		\end{tabular}
	}
	\label{tab:speed_coco}
\end{table}

\begin{table}[!t]
	\centering
	\caption{Results on ADE20k-Full (847 classes) achieved with the GZS3 setting. The FPS is tested on images with the short side of 512 with a single GeForce RTX 3090.}
	\vspace{-3mm}
	\resizebox{0.87\linewidth}{!}{
		\begin{tabular}{c|ccc|c}
			\toprule
			methods       & seen & unseen & harmonic & FPS  \\ \hline
			SPNet-FPN     & 9.2  & 0.9    & 1.6      & 7.9  \\ \hline
			ZegFormer-seg & 18.9 & 1.3    & 2.4      & \textbf{31.3} \\
			ZegFormer     & \textbf{19.7} & \textbf{5.6}    & \textbf{8.7}      & 6.3 \\
			\bottomrule
		\end{tabular}
	}
	\label{tab:speed_ade}
\end{table}

\subsection{Comparisons with Different Backbones.}
Since the ZegFormer-seg with R-50 is better than SPNet-FPN with \textit{R-50} in the supervised semantic segmentation, we also report the results of SPNet-FPN with \textit{R-101}. From Tab.~\ref{tab:backbones}, we can see that the SPNet-FPN with R-101 is comparable with ZegFormer-seg in the supervised evaluation, but much lower than ZegFormer-seg with the GZS3.

\begin{table}[!t]
	\caption{Comparisons with different backbones. In the supervised evaluation, only the pixels of seen classes are evaluated, while the pixels of unseen classes are ignored. The generalized zero-shot evaluation is GZS3, which has been introduced in Sec.~3.1. We can see that SPNet-FPN with \textit{R-101} is comparable with ZegFormer-seg \textit{R-50} in the supervised semantic segmentation, but much lower than ZegFormer-seg with R-50 in the GZS3 evaluation. {\em S}: seen, {\em U}: unseen, and {\em H}: harmonic. }
	\vspace{-3mm}
	\resizebox{\linewidth}{!}{
		\begin{tabular}{c|c|c|ccc|c}
			\toprule
			&          & Supervised & \multicolumn{3}{c|}{GZS3} &      \\ \hline
			method           & backbone & {\em S}       & {\em S}        & {\em U}        & {\em H}       & FPS  \\ \hline
			SPNet-FPN        & R-50     & 38.5       & 32.3       & 11.0         & 16.4          & 17.0 \\
			SPNet-FPN        & R-101    & \textbf{40.7}       & 34.6       & 11.6         & 17.3          & 16.2 \\ \hline
			ZegFormer-seg    & R-50     & 40.3       & \textbf{37.4} & \textbf{21.4}   & \textbf{27.2}     & \textbf{25.5} \\
			\bottomrule
		\end{tabular} \label{tab:backbones}
	}
\end{table}

\section{More Visualization Results}
\begin{figure*}[!t]
	\centering
	\vspace{-3mm}
	\includegraphics[width=0.92\linewidth]{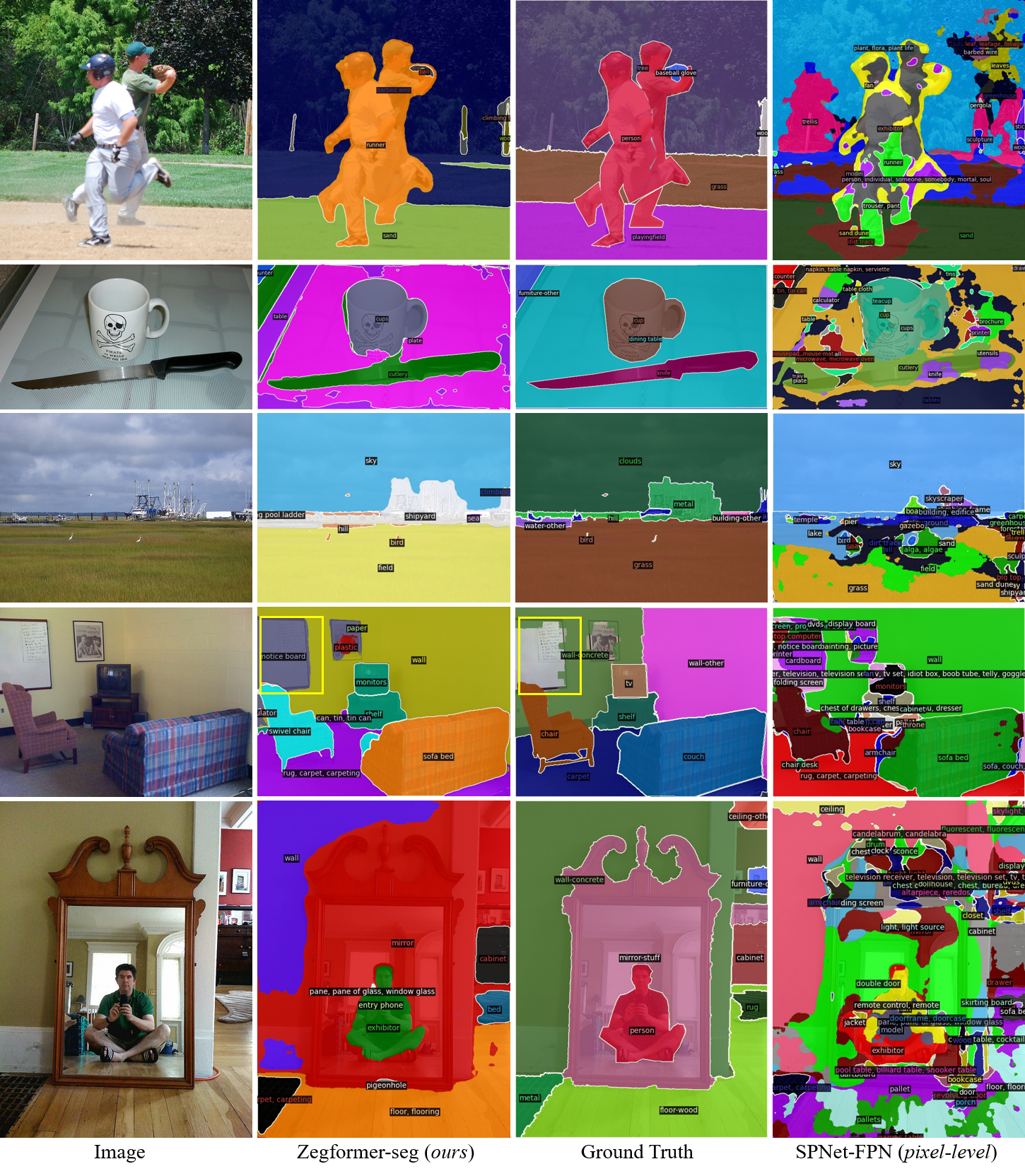}
	\vspace{-3mm}	
	\caption{Results on COCO-Stuff. ZegFormer-seg is our proposed model as an implementation of \textit{decoupling formulation of ZS3}, while the SPNet-FPN is a \textit{pixel-level zero-shot classification} baseline. Both the two models are trained with only 156 classes on COCO-Stuff, and required to segment with 847 class names. We can see that the \textit{pixel-level zero-shot classification} is totally failed when there is a large number of unseen classes. In the yellow box are the unannotated category in COCO-Stuff but segmented by our model.
	}
	\label{fig:adeinfercoco}
\end{figure*}

\begin{figure*}[!t]
	\centering
	\vspace{-3mm}
	\includegraphics[width=0.93\linewidth]{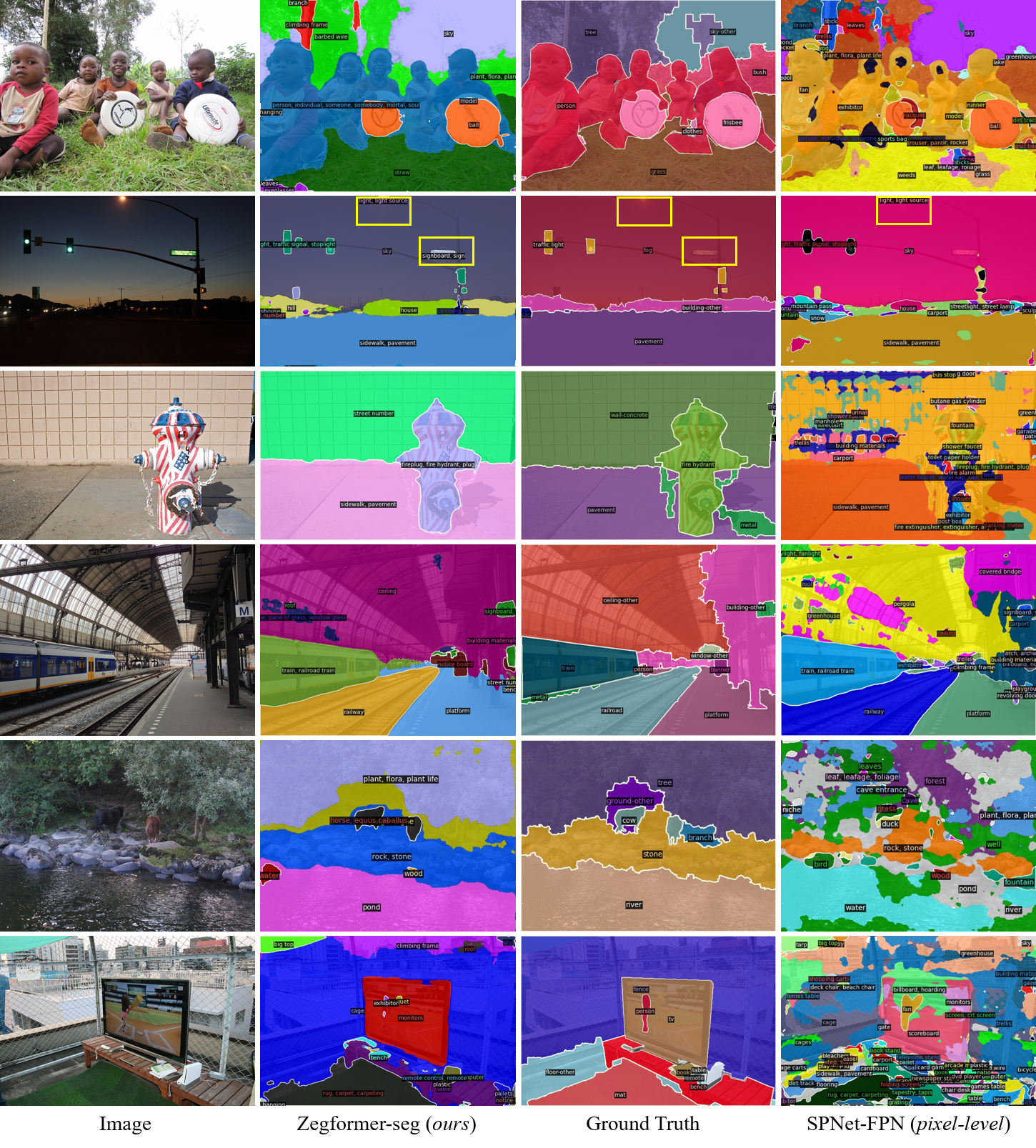}
	\vspace{-3mm}	
	\caption{Results on COCO-Stuff. ZegFormer-seg is our proposed model as an implementation of \textit{decoupling formulation of ZS3}, while the SPNet-FPN is a \textit{pixel-level zero-shot classification} baseline. Both models are trained with only 156 classes on COCO-Stuff, and required to segment with 847 class names. In the yellow box are the unannotated category in COCO-Stuff but segmented by our model.
	}
	\label{fig:adeinfercoco2}
\end{figure*}

We visualize the results of ZegFormer-Seg with R-50 and SPNet-FPN with R-50. The two models are \textit{trained} on COCO-Stuff with \textbf{156 classes}, and required to segment \textbf{847 classes}. The visualization results are shown in Fig.~\ref{fig:adeinfercoco}, and Fig.~\ref{fig:adeinfercoco2}. 

\section{More Implementation Details}
\subsection{HyperParameters}
Following~\cite{maskformer}, we use a FPN~\cite{FPN} structure as the pixel decoder of ZegFormer and SPNet-FPN. The output stride of the pixel decoder is 4.  
Following~\cite{maskformer,detr}, we use 6 Transformer decoder layers and apply the same loss after each layer. The \textit{mask projection} layer in ZegFormer consists of 2 hidden layers of 256 channels. 
During training, we crop images from the original images. The sizes of cropped images are $640\times 640$ in COCO-Stuff, and $512\times 512$ in the ADE20k-Full~\cite{ade20k} and PASCAL VOC~\cite{voc}. 
During testing, we keep the aspect ratio and resize the short size of an image to $640$ in COCO-Stuff, and $512$ in the ADE20k-Full and PASCAL VOC.

\subsection{Prompt Templates}
Following the previous works~\cite{ViLD,clip}, for each category, we used multiple prompt templates to generate the text embeddings then ensemble these text embeddings by averaging. The following is the prompt templates that we used in ZegFormer:
\begin{small}
	\begin{verbatim}
		'a photo of a {}.',
		'This is a photo of a {}',
		'This is a photo of a small {}',
		'This is a photo of a medium {}',
		'This is a photo of a large {}',
		'This is a photo of a {}',
		'This is a photo of a small {}',
		'This is a photo of a medium {}',
		'This is a photo of a large {}',
		'a photo of a {} in the scene',
		'a photo of a {} in the scene',
		'There is a {} in the scene',
		'There is the {} in the scene',
		'This is a {} in the scene',
		'This is the {} in the scene',
		'This is one {} in the scene',
	\end{verbatim}
\end{small}
\begin{table}[]
	\caption{Influence of prompt ensemble.}
	\centering
	\vspace{-3mm}
	\resizebox{\linewidth}{!}{
		\begin{tabular}{c|c|ccc}
			\toprule
			method    & prompt ensemble  & seen & unseen & harmonic \\ \hline
			ZegFormer & \xmark & 35.4 & 32.7   & 34.0     \\
			ZegFormer & \cmark  & \textbf{35.9} & \textbf{33.1}   & \textbf{34.4}    \\
			\bottomrule
		\end{tabular}\label{tab:prompt_ensemble}
	}
\end{table}
To see the influence of prompt templates ensemble, we set a baseline by using only one prompt template, (\textit{i.e.}, ``A photo of the $\{\}$ in the scene.") The comparisons are shown in Tab.~\ref{tab:prompt_ensemble}. We can see that the prompt ensemble will slightly improve the performance.

{\small
\bibliographystyle{ieee_fullname}
\bibliography{egbib}

\begin{thebibliography}{10}\itemsep=-1pt

\bibitem{akata2015evaluation}
Zeynep Akata, Scott Reed, Daniel Walter, Honglak Lee, and Bernt Schiele.
\newblock Evaluation of output embeddings for fine-grained image
  classification.
\newblock In {\em Proceedings of the IEEE conference on computer vision and
  pattern recognition}, pages 2927--2936, 2015.

\bibitem{al2016recovering}
Ziad Al-Halah, Makarand Tapaswi, and Rainer Stiefelhagen.
\newblock Recovering the missing link: Predicting class-attribute associations
  for unsupervised zero-shot learning.
\newblock In {\em Proceedings of the IEEE Conference on Computer Vision and
  Pattern Recognition}, pages 5975--5984, 2016.

\bibitem{baek2021exploiting}
Donghyeon Baek, Youngmin Oh, and Bumsub Ham.
\newblock Exploiting a joint embedding space for generalized zero-shot semantic
  segmentation.
\newblock {\em arXiv preprint arXiv:2108.06536}, 2021.

\bibitem{bucher2017generating}
Maxime Bucher, St{\'e}phane Herbin, and Fr{\'e}d{\'e}ric Jurie.
\newblock Generating visual representations for zero-shot classification.
\newblock In {\em Proceedings of the IEEE International Conference on Computer
  Vision Workshops}, pages 2666--2673, 2017.

\bibitem{ZS3Net}
Maxime Bucher, Tuan-Hung Vu, Matthieu Cord, and Patrick P{\'e}rez.
\newblock Zero-shot semantic segmentation.
\newblock {\em Advances in Neural Information Processing Systems}, 32:468--479,
  2019.

\bibitem{coco-stuff}
Holger Caesar, Jasper Uijlings, and Vittorio Ferrari.
\newblock Coco-stuff: Thing and stuff classes in context.
\newblock In {\em Proceedings of the IEEE conference on computer vision and
  pattern recognition}, pages 1209--1218, 2018.

\bibitem{detr}
Nicolas Carion, Francisco Massa, Gabriel Synnaeve, Nicolas Usunier, Alexander
  Kirillov, and Sergey Zagoruyko.
\newblock End-to-end object detection with transformers.
\newblock In {\em European Conference on Computer Vision}, pages 213--229.
  Springer, 2020.

\bibitem{Deeplab}
Liang{-}Chieh Chen, George Papandreou, Iasonas Kokkinos, Kevin Murphy, and
  Alan~L. Yuille.
\newblock Deeplab: Semantic image segmentation with deep convolutional nets,
  atrous convolution, and fully connected crfs.
\newblock {\em {IEEE} Trans. Pattern Anal. Mach. Intell.}, 40(4):834--848,
  2018.

\bibitem{chen2017deeplab}
Liang-Chieh Chen, George Papandreou, Iasonas Kokkinos, Kevin Murphy, and Alan~L
  Yuille.
\newblock Deeplab: Semantic image segmentation with deep convolutional nets,
  atrous convolution, and fully connected crfs.
\newblock {\em IEEE transactions on pattern analysis and machine intelligence},
  40(4):834--848, 2017.

\bibitem{Deeplabv3Plus}
Liang-Chieh Chen, Yukun Zhu, George Papandreou, Florian Schroff, and Hartwig
  Adam.
\newblock Encoder-decoder with atrous separable convolution for semantic image
  segmentation.
\newblock In {\em Proceedings of the European conference on computer vision
  (ECCV)}, pages 801--818, 2018.

\bibitem{maskformer}
Bowen Cheng, Alexander~G Schwing, and Alexander Kirillov.
\newblock Per-pixel classification is not all you need for semantic
  segmentation.
\newblock {\em arXiv preprint arXiv:2107.06278}, 2021.

\bibitem{SIGN}
Jiaxin Cheng, Soumyaroop Nandi, Prem Natarajan, and Wael Abd-Almageed.
\newblock Sign: Spatial-information incorporated generative network for
  generalized zero-shot semantic segmentation.
\newblock {\em arXiv preprint arXiv:2108.12517}, 2021.

\bibitem{fergus2010semantic}
Rob Fergus, Hector Bernal, Yair Weiss, and Antonio Torralba.
\newblock Semantic label sharing for learning with many categories.
\newblock In {\em European Conference on Computer Vision}, pages 762--775.
  Springer, 2010.

\bibitem{Devise}
Andrea Frome, Greg Corrado, Jonathon Shlens, Samy Bengio, Jeffrey Dean,
  Marc’Aurelio Ranzato, and Tomas Mikolov.
\newblock Devise: A deep visual-semantic embedding model.
\newblock 2013.

\bibitem{fu2017recent}
Yanwei Fu, Tao Xiang, Yu-Gang Jiang, Xiangyang Xue, Leonid Sigal, and Shaogang
  Gong.
\newblock Recent advances in zero-shot recognition.
\newblock {\em arXiv preprint arXiv:1710.04837}, 2017.

\bibitem{ViLD}
Xiuye Gu, Tsung-Yi Lin, Weicheng Kuo, and Yin Cui.
\newblock Zero-shot detection via vision and language knowledge distillation.
\newblock {\em arXiv preprint arXiv:2104.13921}, 2021.

\bibitem{CaGNet}
Zhangxuan Gu, Siyuan Zhou, Li Niu, Zihan Zhao, and Liqing Zhang.
\newblock Context-aware feature generation for zero-shot semantic segmentation.
\newblock In {\em Proceedings of the 28th ACM International Conference on
  Multimedia}, pages 1921--1929, 2020.

\bibitem{CaGNetv2}
Zhangxuan Gu, Siyuan Zhou, Li Niu, Zihan Zhao, and Liqing Zhang.
\newblock From pixel to patch: Synthesize context-aware features for zero-shot
  semantic segmentation.
\newblock {\em arXiv preprint arXiv:2009.12232}, 2020.

\bibitem{maskrcnn}
Kaiming He, Georgia Gkioxari, Piotr Doll{\'a}r, and Ross Girshick.
\newblock Mask r-cnn.
\newblock In {\em Proceedings of the IEEE international conference on computer
  vision}, pages 2961--2969, 2017.

\bibitem{hu2020uncertainty}
Ping Hu, Stan Sclaroff, and Kate Saenko.
\newblock Uncertainty-aware learning for zero-shot semantic segmentation.
\newblock In {\em NeurIPS}, 2020.

\bibitem{jayaraman2014zero}
Dinesh Jayaraman and Kristen Grauman.
\newblock Zero shot recognition with unreliable attributes.
\newblock {\em arXiv preprint arXiv:1409.4327}, 2014.

\bibitem{ji2018end}
Jingwei Ji, Shyamal Buch, Alvaro Soto, and Juan~Carlos Niebles.
\newblock End-to-end joint semantic segmentation of actors and actions in
  video.
\newblock In {\em Proceedings of the European Conference on Computer Vision
  (ECCV)}, pages 702--717, 2018.

\bibitem{kankuekul2012online}
Pichai Kankuekul, Aram Kawewong, Sirinart Tangruamsub, and Osamu Hasegawa.
\newblock Online incremental attribute-based zero-shot learning.
\newblock In {\em 2012 IEEE conference on computer vision and pattern
  recognition}, pages 3657--3664. IEEE, 2012.

\bibitem{kendall2017uncertainties}
Alex Kendall and Yarin Gal.
\newblock What uncertainties do we need in bayesian deep learning for computer
  vision?
\newblock {\em arXiv preprint arXiv:1703.04977}, 2017.

\bibitem{panopticseg}
Alexander Kirillov, Kaiming He, Ross Girshick, Carsten Rother, and Piotr
  Doll{\'a}r.
\newblock Panoptic segmentation.
\newblock In {\em Proceedings of the IEEE/CVF Conference on Computer Vision and
  Pattern Recognition}, pages 9404--9413, 2019.

\bibitem{lampert2013attribute}
Christoph~H Lampert, Hannes Nickisch, and Stefan Harmeling.
\newblock Attribute-based classification for zero-shot visual object
  categorization.
\newblock {\em IEEE transactions on pattern analysis and machine intelligence},
  36(3):453--465, 2013.

\bibitem{CSRL}
Peike Li, Yunchao Wei, and Yi Yang.
\newblock Consistent structural relation learning for zero-shot segmentation.
\newblock {\em Advances in Neural Information Processing Systems}, 33, 2020.

\bibitem{refinenet}
Guosheng Lin, Anton Milan, Chunhua Shen, and Ian Reid.
\newblock Refinenet: Multi-path refinement networks for high-resolution
  semantic segmentation.
\newblock In {\em Proceedings of the IEEE conference on computer vision and
  pattern recognition}, pages 1925--1934, 2017.

\bibitem{FPN}
Tsung-Yi Lin, Piotr Doll{\'a}r, Ross Girshick, Kaiming He, Bharath Hariharan,
  and Serge Belongie.
\newblock Feature pyramid networks for object detection.
\newblock In {\em Proceedings of the IEEE conference on computer vision and
  pattern recognition}, pages 2117--2125, 2017.

\bibitem{FCN}
Jonathan Long, Evan Shelhamer, and Trevor Darrell.
\newblock Fully convolutional networks for semantic segmentation.
\newblock In {\em Proceedings of the IEEE conference on computer vision and
  pattern recognition}, pages 3431--3440, 2015.

\bibitem{long2017zero}
Yang Long, Li Liu, Ling Shao, Fumin Shen, Guiguang Ding, and Jungong Han.
\newblock From zero-shot learning to conventional supervised classification:
  Unseen visual data synthesis.
\newblock In {\em Proceedings of the IEEE Conference on Computer Vision and
  Pattern Recognition}, pages 1627--1636, 2017.

\bibitem{mensink2014costa}
Thomas Mensink, Efstratios Gavves, and Cees~GM Snoek.
\newblock Costa: Co-occurrence statistics for zero-shot classification.
\newblock In {\em Proceedings of the IEEE conference on computer vision and
  pattern recognition}, pages 2441--2448, 2014.

\bibitem{miller1995wordnet}
George~A Miller.
\newblock Wordnet: a lexical database for english.
\newblock {\em Communications of the ACM}, 38(11):39--41, 1995.

\bibitem{norouzi2013zero}
Mohammad Norouzi, Tomas Mikolov, Samy Bengio, Yoram Singer, Jonathon Shlens,
  Andrea Frome, Greg~S Corrado, and Jeffrey Dean.
\newblock Zero-shot learning by convex combination of semantic embeddings.
\newblock {\em arXiv preprint arXiv:1312.5650}, 2013.

\bibitem{STRICT}
Giuseppe Pastore, Fabio Cermelli, Yongqin Xian, Massimiliano Mancini, Zeynep
  Akata, and Barbara Caputo.
\newblock A closer look at self-training for zero-label semantic segmentation.
\newblock In {\em Proceedings of the IEEE/CVF Conference on Computer Vision and
  Pattern Recognition}, pages 2693--2702, 2021.

\bibitem{entityseg}
Lu Qi, Jason Kuen, Yi Wang, Jiuxiang Gu, Hengshuang Zhao, Zhe Lin, Philip Torr,
  and Jiaya Jia.
\newblock Open-world entity segmentation.
\newblock {\em arXiv preprint arXiv:2107.14228}, 2021.

\bibitem{clip}
Alec Radford, Jong~Wook Kim, Chris Hallacy, Aditya Ramesh, Gabriel Goh,
  Sandhini Agarwal, Girish Sastry, Amanda Askell, Pamela Mishkin, Jack Clark,
  et~al.
\newblock Learning transferable visual models from natural language
  supervision.
\newblock {\em arXiv preprint arXiv:2103.00020}, 2021.

\bibitem{rohrbach2011evaluating}
Marcus Rohrbach, Michael Stark, and Bernt Schiele.
\newblock Evaluating knowledge transfer and zero-shot learning in a large-scale
  setting.
\newblock In {\em CVPR 2011}, pages 1641--1648. IEEE, 2011.

\bibitem{rohrbach2010helps}
Marcus Rohrbach, Michael Stark, Gy{\"o}rgy Szarvas, Iryna Gurevych, and Bernt
  Schiele.
\newblock What helps where--and why? semantic relatedness for knowledge
  transfer.
\newblock In {\em 2010 IEEE Computer Society Conference on Computer Vision and
  Pattern Recognition}, pages 910--917. IEEE, 2010.

\bibitem{unet}
Olaf Ronneberger, Philipp Fischer, and Thomas Brox.
\newblock U-net: Convolutional networks for biomedical image segmentation.
\newblock In {\em International Conference on Medical image computing and
  computer-assisted intervention}, pages 234--241. Springer, 2015.

\bibitem{shen2021conterfactual}
Feihong Shen, Jun Liu, and Ping Hu.
\newblock Conterfactual generative zero-shot semantic segmentation.
\newblock {\em arXiv preprint arXiv:2106.06360}, 2021.

\bibitem{segmenter}
Robin Strudel, Ricardo Garcia, Ivan Laptev, and Cordelia Schmid.
\newblock Segmenter: Transformer for semantic segmentation.
\newblock {\em arXiv preprint arXiv:2105.05633}, 2021.

\bibitem{cap2seg}
Guiyu Tian, Shuai Wang, Jie Feng, Li Zhou, and Yadong Mu.
\newblock Cap2seg: Inferring semantic and spatial context from captions for
  zero-shot image segmentation.
\newblock In {\em Proceedings of the 28th ACM International Conference on
  Multimedia}, pages 4125--4134, 2020.

\bibitem{hrnet}
Jingdong Wang, Ke Sun, Tianheng Cheng, Borui Jiang, Chaorui Deng, Yang Zhao,
  Dong Liu, Yadong Mu, Mingkui Tan, Xinggang Wang, et~al.
\newblock Deep high-resolution representation learning for visual recognition.
\newblock {\em IEEE transactions on pattern analysis and machine intelligence},
  2020.

\bibitem{spnet}
Yongqin Xian, Subhabrata Choudhury, Yang He, Bernt Schiele, and Zeynep Akata.
\newblock Semantic projection network for zero-and few-label semantic
  segmentation.
\newblock In {\em Proceedings of the IEEE/CVF Conference on Computer Vision and
  Pattern Recognition}, pages 8256--8265, 2019.

\bibitem{xian2018zero}
Yongqin Xian, Christoph~H Lampert, Bernt Schiele, and Zeynep Akata.
\newblock Zero-shot learning—a comprehensive evaluation of the good, the bad
  and the ugly.
\newblock {\em IEEE transactions on pattern analysis and machine intelligence},
  41(9):2251--2265, 2018.

\bibitem{xian2018feature}
Yongqin Xian, Tobias Lorenz, Bernt Schiele, and Zeynep Akata.
\newblock Feature generating networks for zero-shot learning.
\newblock In {\em Proceedings of the IEEE conference on computer vision and
  pattern recognition}, pages 5542--5551, 2018.

\bibitem{segformer}
Enze Xie, Wenhai Wang, Zhiding Yu, Anima Anandkumar, Jose~M Alvarez, and Ping
  Luo.
\newblock Segformer: Simple and efficient design for semantic segmentation with
  transformers.
\newblock {\em arXiv preprint arXiv:2105.15203}, 2021.

\bibitem{dilatednetwork}
Fisher Yu and Vladlen Koltun.
\newblock Multi-scale context aggregation by dilated convolutions.
\newblock {\em arXiv preprint arXiv:1511.07122}, 2015.

\bibitem{ocrnet}
Yuhui Yuan, Xilin Chen, and Jingdong Wang.
\newblock Object-contextual representations for semantic segmentation.
\newblock In {\em Computer Vision--ECCV 2020: 16th European Conference,
  Glasgow, UK, August 23--28, 2020, Proceedings, Part VI 16}, pages 173--190.
  Springer, 2020.

\bibitem{zhang2016zero}
Ziming Zhang and Venkatesh Saligrama.
\newblock Zero-shot learning via joint latent similarity embedding.
\newblock In {\em proceedings of the IEEE Conference on Computer Vision and
  Pattern Recognition}, pages 6034--6042, 2016.

\bibitem{zhao2017open}
Hang Zhao, Xavier Puig, Bolei Zhou, Sanja Fidler, and Antonio Torralba.
\newblock Open vocabulary scene parsing.
\newblock In {\em Proceedings of the IEEE International Conference on Computer
  Vision}, pages 2002--2010, 2017.

\bibitem{pspnet}
Hengshuang Zhao, Jianping Shi, Xiaojuan Qi, Xiaogang Wang, and Jiaya Jia.
\newblock Pyramid scene parsing network.
\newblock In {\em Proceedings of the IEEE conference on computer vision and
  pattern recognition}, pages 2881--2890, 2017.

\end{thebibliography}


\begin{thebibliography}{10}\itemsep=-1pt

\bibitem{akata2015evaluation}
Zeynep Akata, Scott Reed, Daniel Walter, Honglak Lee, and Bernt Schiele.
\newblock Evaluation of output embeddings for fine-grained image
  classification.
\newblock In {\em CVPR}, pages 2927--2936, 2015.

\bibitem{al2016recovering}
Ziad Al-Halah, Makarand Tapaswi, and Rainer Stiefelhagen.
\newblock Recovering the missing link: Predicting class-attribute associations
  for unsupervised zero-shot learning.
\newblock In {\em CVPR}, pages 5975--5984, 2016.

\bibitem{UCM}
Pablo Arbelaez, Michael Maire, Charless Fowlkes, and Jitendra Malik.
\newblock Contour detection and hierarchical image segmentation.
\newblock {\em IEEE TPAMI}, 33(5):898--916, 2010.

\bibitem{MCG}
Pablo Arbel{\'a}ez, Jordi Pont-Tuset, Jonathan~T Barron, Ferran Marques, and
  Jitendra Malik.
\newblock Multiscale combinatorial grouping.
\newblock In {\em CVPR}, pages 328--335, 2014.

\bibitem{baek2021exploiting}
Donghyeon Baek, Youngmin Oh, and Bumsub Ham.
\newblock Exploiting a joint embedding space for generalized zero-shot semantic
  segmentation.
\newblock {\em arXiv:2108.06536}, 2021.

\bibitem{biederman1987recognition}
Irving Biederman.
\newblock Recognition-by-components: a theory of human image understanding.
\newblock {\em Psychological review}, 94(2):115, 1987.

\bibitem{ZS3Net}
Maxime Bucher, Tuan-Hung Vu, Matthieu Cord, and Patrick P{\'e}rez.
\newblock Zero-shot semantic segmentation.
\newblock {\em NeurIPS}, 32:468--479, 2019.

\bibitem{coco-stuff}
Holger Caesar, Jasper Uijlings, and Vittorio Ferrari.
\newblock Coco-stuff: Thing and stuff classes in context.
\newblock In {\em CVPR}, pages 1209--1218, 2018.

\bibitem{detr}
Nicolas Carion, Francisco Massa, Gabriel Synnaeve, Nicolas Usunier, Alexander
  Kirillov, and Sergey Zagoruyko.
\newblock End-to-end object detection with transformers.
\newblock In {\em ECCV}, pages 213--229. Springer, 2020.

\bibitem{Deeplab}
Liang{-}Chieh Chen, George Papandreou, Iasonas Kokkinos, Kevin Murphy, and
  Alan~L. Yuille.
\newblock Deeplab: Semantic image segmentation with deep convolutional nets,
  atrous convolution, and fully connected crfs.
\newblock {\em IEEE TPAMI}, 40(4):834--848, 2018.

\bibitem{Deeplabv3Plus}
Liang-Chieh Chen, Yukun Zhu, George Papandreou, Florian Schroff, and Hartwig
  Adam.
\newblock Encoder-decoder with atrous separable convolution for semantic image
  segmentation.
\newblock In {\em ECCV}, pages 801--818, 2018.

\bibitem{maskformer}
Bowen Cheng, Alexander~G Schwing, and Alexander Kirillov.
\newblock Per-pixel classification is not all you need for semantic
  segmentation.
\newblock {\em NeurIPS}, 2021.

\bibitem{SIGN}
Jiaxin Cheng, Soumyaroop Nandi, Prem Natarajan, and Wael Abd-Almageed.
\newblock Sign: Spatial-information incorporated generative network for
  generalized zero-shot semantic segmentation.
\newblock {\em arXiv:2108.12517}, 2021.

\bibitem{deng1999color}
Yining Deng, B~Shin Manjunath, and Hyundoo Shin.
\newblock Color image segmentation.
\newblock In {\em CVPR}, volume~2, pages 446--451. IEEE, 1999.

\bibitem{voc}
Mark Everingham, Luc Van~Gool, Christopher~KI Williams, John Winn, and Andrew
  Zisserman.
\newblock The pascal visual object classes (voc) challenge.
\newblock {\em IJCV}, 88(2):303--338, 2010.

\bibitem{fergus2010semantic}
Rob Fergus, Hector Bernal, Yair Weiss, and Antonio Torralba.
\newblock Semantic label sharing for learning with many categories.
\newblock In {\em ECCV}, pages 762--775. Springer, 2010.

\bibitem{Devise}
Andrea Frome, Gregory~S. Corrado, Jonathon Shlens, Samy Bengio, Jeffrey Dean,
  Marc'Aurelio Ranzato, and Tom{\'{a}}s Mikolov.
\newblock Devise: {A} deep visual-semantic embedding model.
\newblock In {\em NeurIPS}, 2013.

\bibitem{fu2017recent}
Yanwei Fu, Tao Xiang, Yu-Gang Jiang, Xiangyang Xue, Leonid Sigal, and Shaogang
  Gong.
\newblock Recent advances in zero-shot recognition.
\newblock {\em arXiv:1710.04837}, 2017.

\bibitem{ViLD}
Xiuye Gu, Tsung-Yi Lin, Weicheng Kuo, and Yin Cui.
\newblock Zero-shot detection via vision and language knowledge distillation.
\newblock {\em arXiv:2104.13921}, 2021.

\bibitem{CaGNet}
Zhangxuan Gu, Siyuan Zhou, Li Niu, Zihan Zhao, and Liqing Zhang.
\newblock Context-aware feature generation for zero-shot semantic segmentation.
\newblock In {\em ACM MM}, pages 1921--1929, 2020.

\bibitem{CaGNetv2}
Zhangxuan Gu, Siyuan Zhou, Li Niu, Zihan Zhao, and Liqing Zhang.
\newblock From pixel to patch: Synthesize context-aware features for zero-shot
  semantic segmentation.
\newblock {\em arXiv:2009.12232}, 2020.

\bibitem{maskrcnn}
Kaiming He, Georgia Gkioxari, Piotr Doll{\'a}r, and Ross Girshick.
\newblock Mask r-cnn.
\newblock In {\em ICCV}, pages 2961--2969, 2017.

\bibitem{everything}
Ronghang Hu, Piotr Doll{\'a}r, Kaiming He, Trevor Darrell, and Ross Girshick.
\newblock Learning to segment every thing.
\newblock In {\em CVPR}, pages 4233--4241, 2018.

\bibitem{jayaraman2014zero}
Dinesh Jayaraman and Kristen Grauman.
\newblock Zero shot recognition with unreliable attributes.
\newblock {\em arXiv:1409.4327}, 2014.

\bibitem{ji2018end}
Jingwei Ji, Shyamal Buch, Alvaro Soto, and Juan~Carlos Niebles.
\newblock End-to-end joint semantic segmentation of actors and actions in
  video.
\newblock In {\em ECCV}, pages 702--717, 2018.

\bibitem{align}
Chao Jia, Yinfei Yang, Ye Xia, Yi-Ting Chen, Zarana Parekh, Hieu Pham, Quoc~V
  Le, Yunhsuan Sung, Zhen Li, and Tom Duerig.
\newblock Scaling up visual and vision-language representation learning with
  noisy text supervision.
\newblock {\em arXiv:2102.05918}, 2021.

\bibitem{fasttext}
Armand Joulin, Edouard Grave, Piotr Bojanowski, Matthijs Douze, H{\'e}rve
  J{\'e}gou, and Tomas Mikolov.
\newblock Fasttext. zip: Compressing text classification models.
\newblock {\em arXiv:1612.03651}, 2016.

\bibitem{kankuekul2012online}
Pichai Kankuekul, Aram Kawewong, Sirinart Tangruamsub, and Osamu Hasegawa.
\newblock Online incremental attribute-based zero-shot learning.
\newblock In {\em CVPR}, pages 3657--3664. IEEE, 2012.

\bibitem{kendall2017uncertainties}
Alex Kendall and Yarin Gal.
\newblock What uncertainties do we need in bayesian deep learning for computer
  vision?
\newblock {\em arXiv:1703.04977}, 2017.

\bibitem{lampert2013attribute}
Christoph~H Lampert, Hannes Nickisch, and Stefan Harmeling.
\newblock Attribute-based classification for zero-shot visual object
  categorization.
\newblock {\em IEEE TPAMI}, 36(3):453--465, 2013.

\bibitem{CSRL}
Peike Li, Yunchao Wei, and Yi Yang.
\newblock Consistent structural relation learning for zero-shot segmentation.
\newblock {\em NeurIPS}, 33, 2020.

\bibitem{panopticsegformer}
Zhiqi Li, Wenhai Wang, Enze Xie, Zhiding Yu, Anima Anandkumar, Jose~M Alvarez,
  Tong Lu, and Ping Luo.
\newblock Panoptic segformer.
\newblock {\em arXiv:2109.03814}, 2021.

\bibitem{FPN}
Tsung-Yi Lin, Piotr Doll{\'a}r, Ross Girshick, Kaiming He, Bharath Hariharan,
  and Serge Belongie.
\newblock Feature pyramid networks for object detection.
\newblock In {\em CVPR}, pages 2117--2125, 2017.

\bibitem{focalloss}
Tsung-Yi Lin, Priya Goyal, Ross Girshick, Kaiming He, and Piotr Doll{\'a}r.
\newblock Focal loss for dense object detection.
\newblock In {\em ICCV}, pages 2980--2988, 2017.

\bibitem{liu2021efficient}
Yahui Liu, Enver Sangineto, Wei Bi, Nicu Sebe, Bruno Lepri, and Marco De~Nadai.
\newblock Efficient training of visual transformers with small-size datasets.
\newblock {\em arXiv:2106.03746}, 2021.

\bibitem{FCN}
Jonathan Long, Evan Shelhamer, and Trevor Darrell.
\newblock Fully convolutional networks for semantic segmentation.
\newblock In {\em CVPR}, pages 3431--3440, 2015.

\bibitem{mensink2014costa}
Thomas Mensink, Efstratios Gavves, and Cees~GM Snoek.
\newblock Costa: Co-occurrence statistics for zero-shot classification.
\newblock In {\em CVPR}, pages 2441--2448, 2014.

\bibitem{word2vec}
Tomas Mikolov, Ilya Sutskever, Kai Chen, Greg~S Corrado, and Jeff Dean.
\newblock Distributed representations of words and phrases and their
  compositionality.
\newblock In {\em NeurIPS}, pages 3111--3119, 2013.

\bibitem{miller1995wordnet}
George~A Miller.
\newblock Wordnet: a lexical database for english.
\newblock {\em CACM}, 38(11):39--41, 1995.

\bibitem{dice}
Fausto Milletari, Nassir Navab, and Seyed-Ahmad Ahmadi.
\newblock V-net: Fully convolutional neural networks for volumetric medical
  image segmentation.
\newblock In {\em 3DV}, pages 565--571. IEEE, 2016.

\bibitem{norouzi2013zero}
Mohammad Norouzi, Tomas Mikolov, Samy Bengio, Yoram Singer, Jonathon Shlens,
  Andrea Frome, Greg~S Corrado, and Jeffrey Dean.
\newblock Zero-shot learning by convex combination of semantic embeddings.
\newblock {\em arXiv:1312.5650}, 2013.

\bibitem{STRICT}
Giuseppe Pastore, Fabio Cermelli, Yongqin Xian, Massimiliano Mancini, Zeynep
  Akata, and Barbara Caputo.
\newblock A closer look at self-training for zero-label semantic segmentation.
\newblock In {\em CVPRW}, pages 2693--2702, 2021.

\bibitem{deepmask}
Pedro~O Pinheiro, Ronan Collobert, and Piotr Doll{\'a}r.
\newblock Learning to segment object candidates.
\newblock {\em arXiv:1506.06204}, 2015.

\bibitem{supervisedeval}
Jordi Pont-Tuset and Ferran Marques.
\newblock Supervised evaluation of image segmentation and object proposal
  techniques.
\newblock {\em IEEE TPAMI}, 38(7):1465--1478, 2015.

\bibitem{entityseg}
Lu Qi, Jason Kuen, Yi Wang, Jiuxiang Gu, Hengshuang Zhao, Zhe Lin, Philip Torr,
  and Jiaya Jia.
\newblock Open-world entity segmentation.
\newblock {\em arXiv:2107.14228}, 2021.

\bibitem{clip}
Alec Radford, Jong~Wook Kim, Chris Hallacy, Aditya Ramesh, Gabriel Goh,
  Sandhini Agarwal, Girish Sastry, Amanda Askell, Pamela Mishkin, Jack Clark,
  Gretchen Krueger, and Ilya Sutskever.
\newblock Learning transferable visual models from natural language
  supervision.
\newblock In {\em ICML}, pages 8748--8763, 2021.

\bibitem{rohrbach2011evaluating}
Marcus Rohrbach, Michael Stark, and Bernt Schiele.
\newblock Evaluating knowledge transfer and zero-shot learning in a large-scale
  setting.
\newblock In {\em CVPR}, pages 1641--1648. IEEE, 2011.

\bibitem{rohrbach2010helps}
Marcus Rohrbach, Michael Stark, Gy{\"o}rgy Szarvas, Iryna Gurevych, and Bernt
  Schiele.
\newblock What helps where--and why? semantic relatedness for knowledge
  transfer.
\newblock In {\em CVPR}, pages 910--917. IEEE, 2010.

\bibitem{shen2021conterfactual}
Feihong Shen, Jun Liu, and Ping Hu.
\newblock Conterfactual generative zero-shot semantic segmentation.
\newblock {\em arXiv:2106.06360}, 2021.

\bibitem{NormalizeCUT}
Jianbo Shi and Jitendra Malik.
\newblock Normalized cuts and image segmentation.
\newblock {\em IEEE TPAMI}, 22(8):888--905, 2000.

\bibitem{cap2seg}
Guiyu Tian, Shuai Wang, Jie Feng, Li Zhou, and Yadong Mu.
\newblock Cap2seg: Inferring semantic and spatial context from captions for
  zero-shot image segmentation.
\newblock In {\em ACM MM}, pages 4125--4134, 2020.

\bibitem{DDMCMC}
Zhuowen Tu and Song-Chun Zhu.
\newblock Image segmentation by data-driven markov chain monte carlo.
\newblock {\em IEEE TPAMI}, 24(5):657--673, 2002.

\bibitem{wu2019detectron2}
Yuxin Wu, Alexander Kirillov, Francisco Massa, Wan-Yen Lo, and Ross Girshick.
\newblock Detectron2.
\newblock \url{https://github.com/facebookresearch/detectron2}, 2019.

\bibitem{spnet}
Yongqin Xian, Subhabrata Choudhury, Yang He, Bernt Schiele, and Zeynep Akata.
\newblock Semantic projection network for zero-and few-label semantic
  segmentation.
\newblock In {\em CVPR}, pages 8256--8265, 2019.

\bibitem{xian2018zero}
Yongqin Xian, Christoph~H Lampert, Bernt Schiele, and Zeynep Akata.
\newblock Zero-shot learning—a comprehensive evaluation of the good, the bad
  and the ugly.
\newblock {\em IEEE TPAMI}, 41(9):2251--2265, 2018.

\bibitem{segformer}
Enze Xie, Wenhai Wang, Zhiding Yu, Anima Anandkumar, Jose~M Alvarez, and Ping
  Luo.
\newblock Segformer: Simple and efficient design for semantic segmentation with
  transformers.
\newblock {\em NeurIPS}, 2021.

\bibitem{ocrnet}
Yuhui Yuan, Xilin Chen, and Jingdong Wang.
\newblock Object-contextual representations for semantic segmentation.
\newblock In {\em ECCV}, pages 173--190. Springer, 2020.

\bibitem{openvocdet}
Alireza Zareian, Kevin~Dela Rosa, Derek~Hao Hu, and Shih-Fu Chang.
\newblock Open-vocabulary object detection using captions.
\newblock In {\em CVPR}, pages 14393--14402, 2021.

\bibitem{KNet}
Wenwei Zhang, Jiangmiao Pang, Kai Chen, and Chen~Change Loy.
\newblock K-net: Towards unified image segmentation.
\newblock {\em arXiv:2106.14855}, 2021.

\bibitem{zhang2016zero}
Ziming Zhang and Venkatesh Saligrama.
\newblock Zero-shot learning via joint latent similarity embedding.
\newblock In {\em CVPR}, pages 6034--6042, 2016.

\bibitem{zhao2017open}
Hang Zhao, Xavier Puig, Bolei Zhou, Sanja Fidler, and Antonio Torralba.
\newblock Open vocabulary scene parsing.
\newblock In {\em ICCV}, pages 2002--2010, 2017.

\bibitem{pspnet}
Hengshuang Zhao, Jianping Shi, Xiaojuan Qi, Xiaogang Wang, and Jiaya Jia.
\newblock Pyramid scene parsing network.
\newblock In {\em CVPR}, pages 2881--2890, 2017.

\bibitem{zeroshotinst}
Ye Zheng, Jiahong Wu, Yongqiang Qin, Faen Zhang, and Li Cui.
\newblock Zero-shot instance segmentation.
\newblock In {\em CVPR}, pages 2593--2602, 2021.

\bibitem{ade20k}
Bolei Zhou, Hang Zhao, Xavier Puig, Sanja Fidler, Adela Barriuso, and Antonio
  Torralba.
\newblock Scene parsing through ade20k dataset.
\newblock In {\em CVPR}, pages 633--641, 2017.

\bibitem{zhu1996region}
Song~Chun Zhu and Alan Yuille.
\newblock Region competition: Unifying snakes, region growing, and bayes/mdl
  for multiband image segmentation.
\newblock {\em IEEE TPAMI}, 18(9):884--900, 1996.

\end{thebibliography}
}

\end{document}